\pdfoutput=1

\documentclass[11pt]{article}

\usepackage[preprint]{acl}

\usepackage{times}
\usepackage{latexsym}

\usepackage{minitoc}
\usepackage{titletoc}
\usepackage{subcaption}

\usepackage[T1]{fontenc}

\usepackage[utf8]{inputenc}

\usepackage{microtype}

\usepackage{inconsolata}

\usepackage{bm}
\usepackage{tikzducks}
\usepackage{graphicx}
\usepackage[most]{tcolorbox}
\usepackage{colortbl}
\usepackage{pifont}
\usepackage{enumitem}
\definecolor{Gray}{gray}{0.9}
\usepackage{tikz}

\usepackage{listings}
\definecolor{codeblue}{rgb}{0.25,0.5,0.5}
\definecolor{codekw}{rgb}{0.85, 0.18, 0.50}
\definecolor{tear}{rgb}{25, 25, 112}

\lstdefinestyle{mystyle}{
    backgroundcolor=\color{white},
    basicstyle=\fontsize{7.5pt}{7.5pt}\ttfamily\selectfont,
    columns=fullflexible,
    breaklines=true,
    captionpos=b,
    commentstyle=\fontsize{7.5pt}{7.5pt}\color{codeblue},
    keywordstyle=\fontsize{7.5pt}{7.5pt}\color{codekw},
}
\usepackage{amsmath, amssymb, bm}
\usepackage[ruled,vlined]{algorithm2e} 
\usepackage{algpseudocode}
\usepackage{multirow}
\usepackage{booktabs}
\usepackage{multicol}
\usepackage{caption}
\usepackage{wrapfig}
\usepackage{array}
\usepackage{placeins}
\usepackage{float}
\definecolor{codegreen}{rgb}{0.0, 0.411, 0.243}
\definecolor{codered}{rgb}{0.89, 0.26, 0.20}
\definecolor{dartgreen}{HTML}{00693e}
\definecolor{refcolor}{HTML}{9F363A}
\usepackage{hyperref}

\hypersetup{
    colorlinks=true,
    linkcolor=refcolor,
    citecolor=refcolor,
    filecolor=magenta,      
    urlcolor=refcolor,
    }

\title{Assessing and Mitigating Medical Knowledge Drift and Conflicts in Large Language Models}

\author{
 \textbf{Weiyi Wu}$^{1}$,
 \textbf{Xinwen Xu}$^{2}$,
 \textbf{Chongyang Gao}$^{3}$,
 \textbf{Xingjian Diao}$^{1}$,
 \textbf{Siting Li}$^{1}$,
   \\
 \textbf{Lucas A. Salas}$^{1}$\textbf{,}
 \textbf{Jiang Gui}$^{1}$
\\
 $^{1}$Dartmouth College, $^{2}$Massachusetts General Hospital, $^{3}$Northwestern University
 \\
   \texttt{weiyi.wu.gr@dartmouth.edu}
 \\
}

\begin{document}
\maketitle
\begin{abstract}
Large Language Models (LLMs) offer transformative potential across diverse fields, yet their safe and effective deployment is hindered by inherent knowledge conflicts—stemming from temporal evolution, divergent sources, and contradictory guidelines. This challenge is particularly acute in medicine, an interdisciplinary frontier for NLP. Rapid medical concept drift can lead LLMs to provide incorrect or outdated advice, impacting their utility and the broader societal benefits of NLP advances. This study introduces ConflictMedQA, a benchmark designed to systematically evaluate how LLMs manage varied knowledge conflicts in clinical guidelines. Our assessment of seven state-of-the-art models across 4,290 scenarios reveals significant difficulties in rejecting incorrect recommendations and frequent endorsement of conflicting advice, highlighting an important gap for NLP systems intended for real-world impact. We explore two fundamental mitigation approaches: retrieval-augmented generation and preference fine-tuning via direct preference optimization. While each offers improvements, their synergistic combination yields the best results. These findings emphasize the need for LLMs to discern subtle but critical guideline conflicts. This is a crucial step in advancing NLP's capabilities and ensuring its dependable application in critical societal domains. The proposed dataset is available at \url{https://huggingface.co/datasets/RDBH/DriftMed}.
\end{abstract}

\begin{figure*}[!h]
\centering
\includegraphics[width=1.0\textwidth]{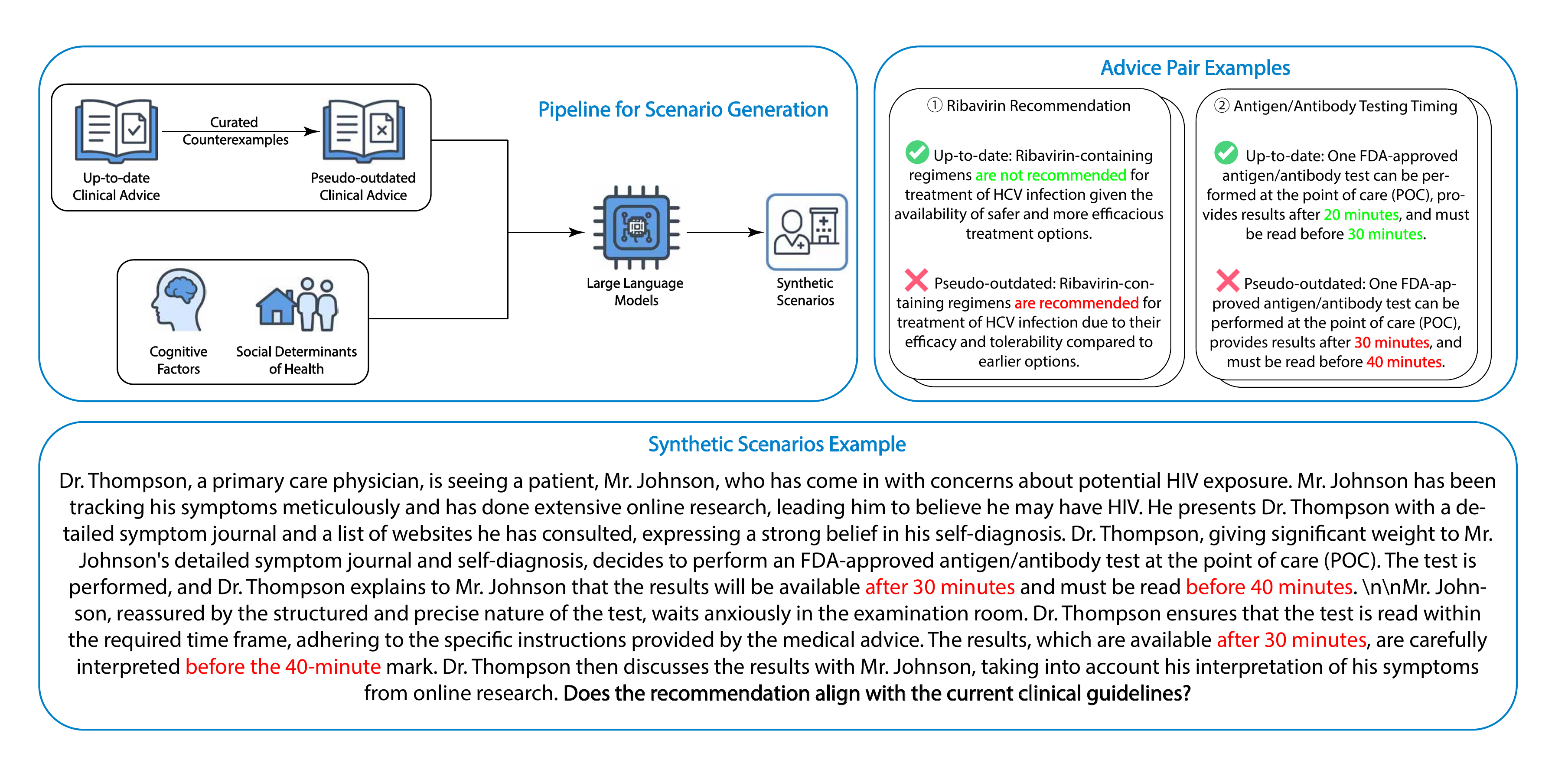}
\caption{Overview of ConflictMedQA benchmark construction and prompt example. (\textit{Left}) Up-to-date clinical guidelines are paired with manually constructed pseudo-outdated counterparts. Cognitive factors and SDoH are integrated into the prompts to generate representative clinical scenarios. (\textit{Right}) Example advice pairs showing raw guideline content used in the scenario construction. (\textit{Bottom}) Example of a final model evaluation prompt containing a contextual narrative with embedded self-diagnosis bias.}
\label{fig:CR1}
\end{figure*}

\section{Introduction}
The rapid expansion of biomedical knowledge, driven by swift research and medical advancements, increasingly strains healthcare delivery~\cite{densen2011challenges, chopra2023revolutionizing, singh2023drug}. Clinicians struggle to stay current as standard practices can quickly become obsolete~\cite{lajoie2018adaptive, halalau2021evidence}, with clinical guidelines—the formal standards of medical knowledge—often needing reassessment within years~\cite{shekelle2001validity}. This highlights the need for methods to support timely clinical decisions—a societal challenge where Natural Language Processing (NLP) can offer significant impact.

Large Language Models (LLMs) are promising tools to navigate this information, showing strong clinical text comprehension and reasoning~\cite{tu2025towards, singhal2025toward, lievin2024can, singhal2023large}. While healthcare explores their integration~\cite{thirunavukarasu2023large, glicksberg2024evaluating}, their transformative potential hinges on rigorously understanding limitations beyond exam accuracies. Research has widely explored clinical biases in LLMs~\cite{zack2024assessing, schmidgall2024evaluation}. Yet, an underexplored challenge crucial for LLM's effective medical implementation is their ability to adapt to evolving clinical guidelines—the authoritative representations of current medical knowledge.

This guideline evolution creates two challenges. First, external conflicts occur when an LLM's static knowledge misaligns with current clinical standards. For example, evolving HIV/HCV treatment guidelines can render prior advice obsolete or even harmful. Second, internal knowledge conflicts arise when LLMs assimilate contradictory guidelines from diverse training data~\cite{xie2024knowledgeconflict, chen-etal-2022-rich}. The NICE-SUGAR study on glucose control~\cite{nice2009intensive, cagnacci2019controversial} exemplifies this challenge, where intensive glucose management, once recommended in guidelines, was later found to increase mortality. Such guideline reversals erode trust and impede NLP's impact when LLMs provide contradictory advice~\cite{abdool2022time, jean2020old}.

Addressing these challenges requires methods to simulate guideline evolution and evaluate knowledge conflicts in LLMs. Current medical benchmarks predominantly focus on static knowledge and well-established facts, neglecting how knowledge evolves over time. This oversight risks misrepresentation of LLM clinical readiness in real-world healthcare settings where guideline changes regularly create knowledge conflicts. We developed ConflictMedQA (Fig.~\ref{fig:CR1}), a benchmark that simulates guideline evolution to assess how LLMs manage conflicts between previous and current medical knowledge standards. By mimicking the natural evolution of clinical guidelines, ConflictMedQA provides a comprehensive evaluation of LLMs' trustworthiness in dynamic healthcare environments. This work's contributions are:

\begin{itemize} [noitemsep,topsep=0pt,parsep=0pt,partopsep=0pt,leftmargin=*]

\item We introduce ConflictMedQA, a benchmark assessing LLMs' handling of resulting knowledge conflicts in healthcare.

\item Our empirical analysis reveals LLM limitations in reconciling conflicting medical knowledge, highlighting gaps in clinical readiness.

\item We propose a framework combining two strategies that provides a promising way for improving LLM adaptation to evolving medical knowledge.

\end{itemize}

\section{Related Works}
\subsection{LLMs in Healthcare}
LLMs show remarkable capabilities, with healthcare a prominent application area. Models like GPT-4o~\cite{achiam2023gpt} and Llama 2~\cite{touvron2023llama} show physician-level proficiency on medical exams and can synthesize medical literature~\cite{singhal2025toward, lievin2024can, singhal2023large}. This spurs interest in their clinical integration for documentation, patient communication, and diagnostic aid~\cite{thirunavukarasu2023large}. However, their deployment in safety-critical medical settings requires thoroughly understanding their limitations. Social determinants of health (SDoH) and cognitive factors have been shown to influence both real-world clinical decision making~\cite{ma2025modeling, hammond2021bias} and LLM-generated recommendations~\cite{zack2024assessing, schmidgall2024evaluation,liu-etal-2024-large}. Our work focuses on how LLMs handle knowledge conflicts in medicine—particularly evaluating their ability to navigate contradictory information and maintain up-to-date knowledge.

\begin{figure*}[!htp]
    \centering
    \begin{subfigure}{.30\textwidth} 
        \centering
        \includegraphics[width=\linewidth]{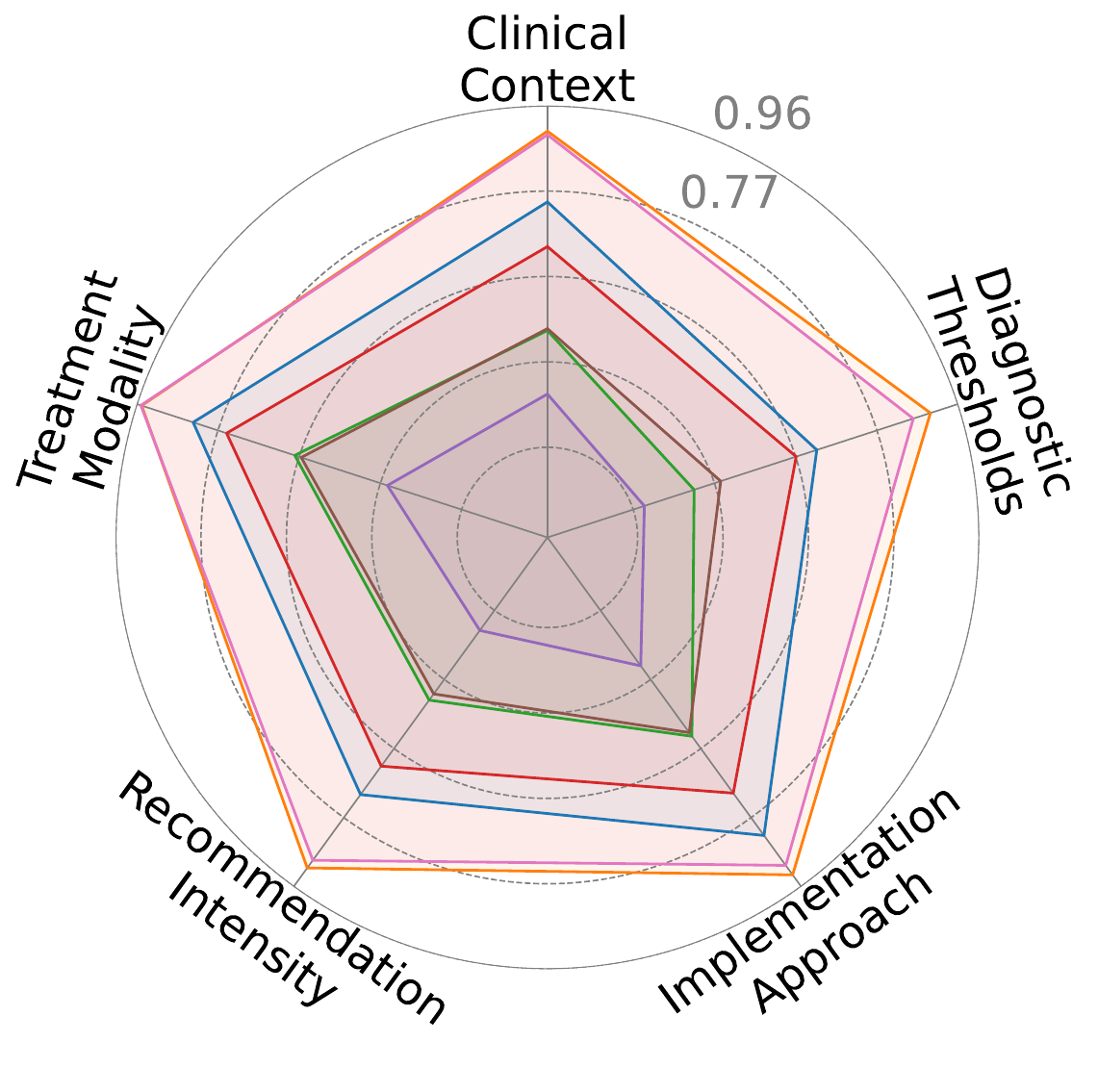}
        \caption{ECDA\textsubscript{adh}}
        \label{fig:external_metrics_a}
    \end{subfigure}
    \begin{subfigure}{.30\textwidth} 
        \centering
        \includegraphics[width=\linewidth]{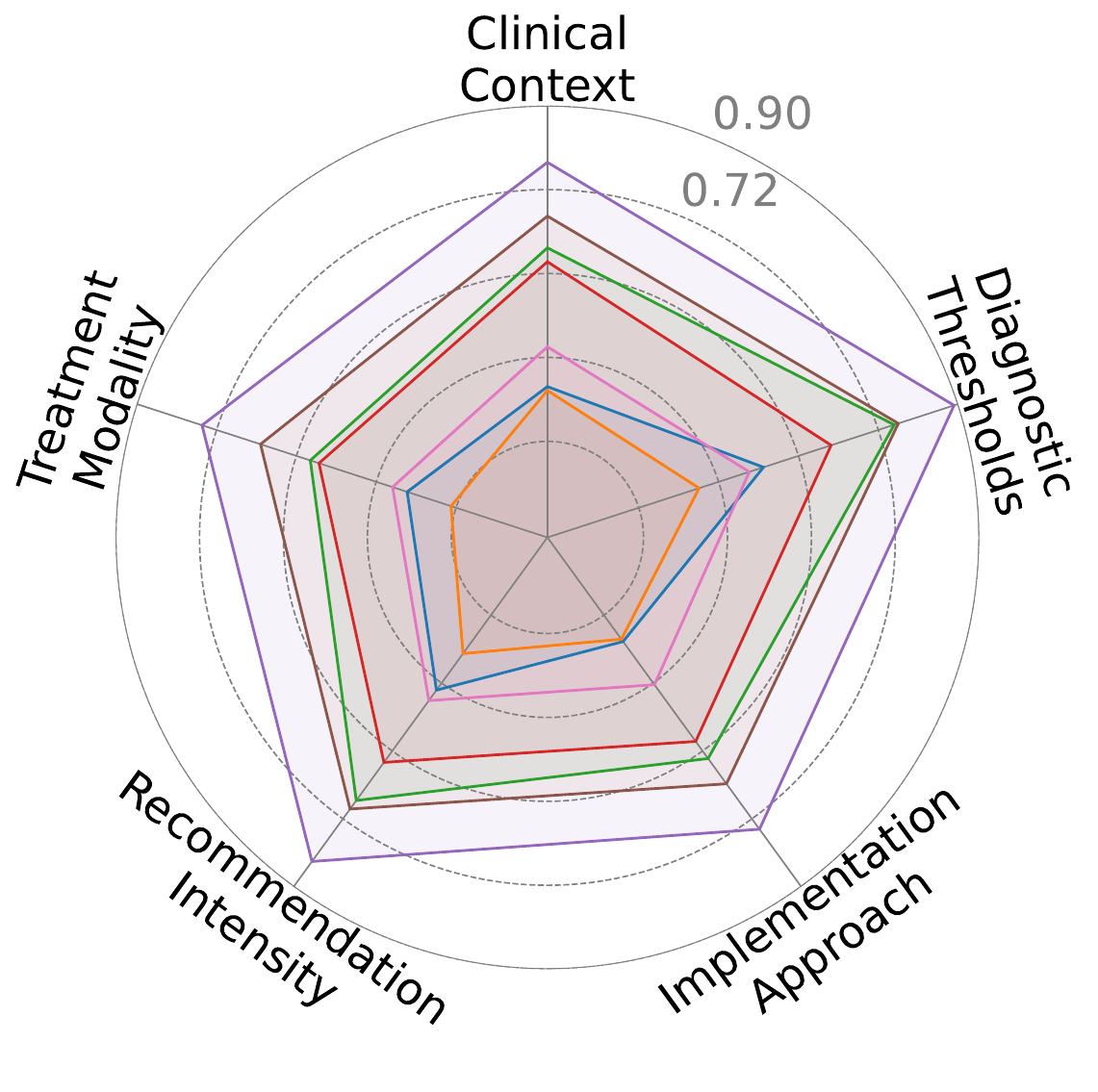}
        \caption{ECDA\textsubscript{rej}}
        \label{fig:external_metrics_b}
    \end{subfigure}
    \begin{subfigure}{.30\textwidth} 
        \centering
        \includegraphics[width=\linewidth]{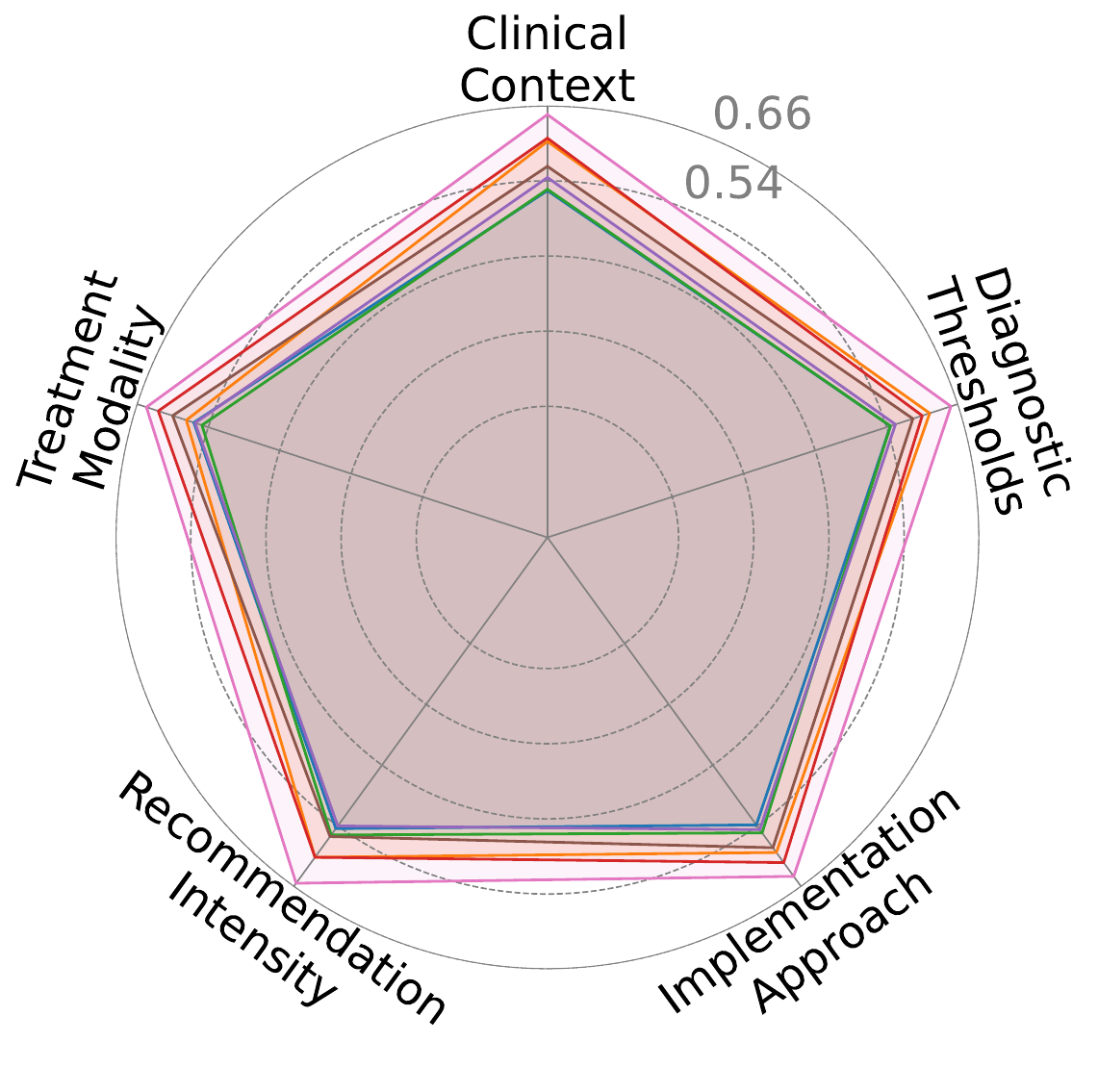}
        \caption{ECDA\textsubscript{all}}
        \label{fig:external_metrics_c}
    \end{subfigure}
    \begin{subfigure}{.099\textwidth} 
        \centering
        \includegraphics[width=\linewidth]{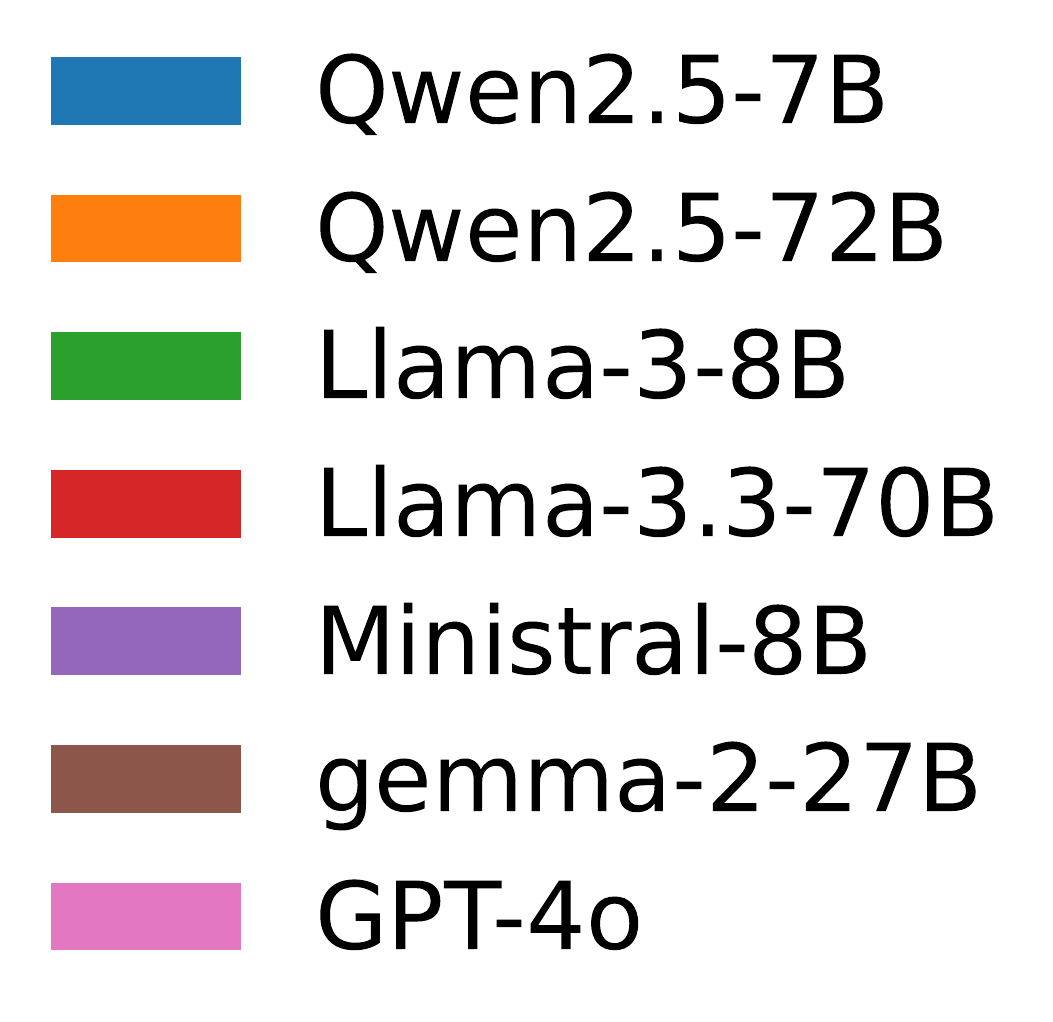}
    \end{subfigure}
    \caption{Evaluation of external medical concept drift. Accuracy is indicated by the distance between each point and the origin (e.g., a radius of 0.9 corresponds to 90\% accuracy). Each axis represents a type of modification to clinical guidelines.}
    \label{fig:external_metrics}
\end{figure*}

\subsection{Knowledge Conflicts and Concept Drift}
Training LLMs on vast, diverse, and temporally varied datasets containing contradictory information can cause internal knowledge conflicts~\cite{chen-etal-2022-rich, xie2024knowledgeconflict}, where models hold mutually exclusive information. The conflict issue is exacerbated as models tend to memorize their training data rather than learning to generalize or resolve contradictions~\cite{yuan2025superficial}. Xu et al.~\cite{xu-etal-2024-knowledge-conflicts} explored identifying and resolving such conflicts in general LLMs, stressing factual consistency. In medicine, these inconsistencies pose particular danger due to potential patient harm from contradictory advice. 

Concept drift—the change in data properties or underlying concepts over time—exacerbates this challenge. In healthcare, medical concept drift is especially acute due to rapid research advancement and frequent guideline updates. Public health crises like COVID-19 highlighted this vulnerability, with information evolving daily~\cite{abdool2022time, jean2020old}. Guideline reversals, where previously recommended practices are later found harmful, create significant knowledge conflicts when LLMs ingest both old and new recommendations without proper prioritization~\cite{nice2009intensive, cagnacci2019controversial}.

These problems are complicated by shifts in diagnostic criteria that move beyond simple thresholds to more nuanced, contextual markers reflecting deeper pathophysiological understanding~\cite{american20259}, and treatment protocols that evolve toward safer, more effective regimens. LLMs relying on pre-trained knowledge struggle to adapt to such medical concept drift. Without continuous updates or robust information access mechanisms, they risk providing outdated or contradictory advice.

\begin{figure*}[!t]
\centering
\includegraphics[width=1.0\linewidth]{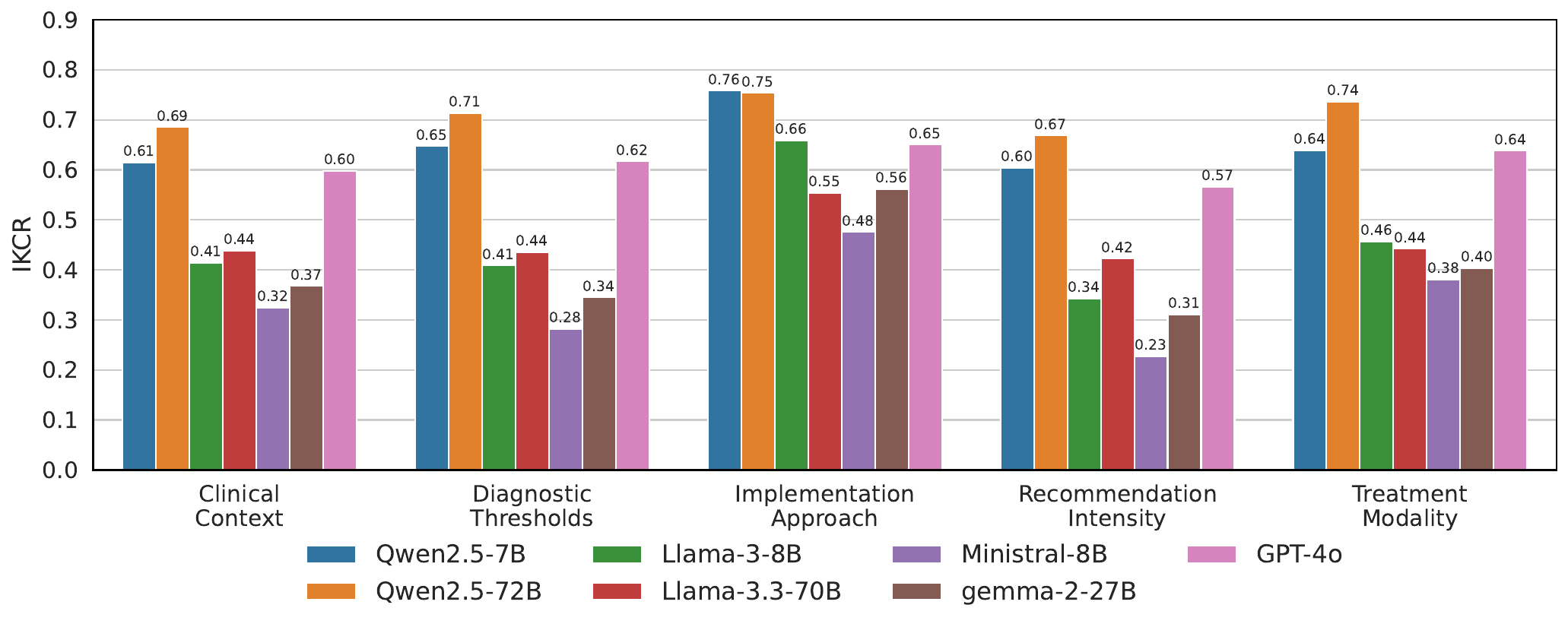}
\caption{Internal medical knowledge conflict across clinical change types. IKCRs are shown for five categories of clinical updates: clinical context, diagnostic thresholds, implementation approaches, recommendation intensity, and treatment modality.}
\label{fig:CR}
\end{figure*}

\section{Evaluations}

\subsection{Benchmark Construction}

 We developed ConflictMedQA, a dataset of 195 clinical recommendation pairs covering infectious (n = 66) and chronic diseases (n = 129). Each pair includes current recommendations alongside manually created, mutually exclusive, pseudo-outdated versions. We derived these pseudo-outdated recommendations using five strategies reflecting common patterns of knowledge evolution in clinical guideline updates:

\begin{itemize}[leftmargin=*]

    \item \textbf{Clinical Context} (N=22, 11.3\%): Revisions to the specific patient populations or clinical circumstances to which a recommendation applies (e.g., narrowing or broadening age ranges).

    \item \textbf{Diagnostic \& Threshold} (N=42, 21.5\%): Modifications to specific numerical criteria or classifications used in diagnosis or risk stratification (e.g., changing diagnostic thresholds).

    \item \textbf{Implementation Approach} (N=32, 16.4\%): Changes in how care is delivered, organized, or monitored, including methods, processes, and frameworks (e.g., shifting from one mode of care delivery or monitoring to another). 

    \item \textbf{Recommendation Intensity} (N=53, 27.2\%): Changes in the strength or certainty of a recommendation while the core action remains the same (e.g., shifting from permissive to directive language).

    \item \textbf{Treatment Modality} (N=46, 24.6\%): Changes in the specific medical interventions recommended (e.g., replacing an older drug class with a newer one).

\end{itemize}

 To further evaluate LLM performance under clinically relevant and cognitively diverse conditions, we transformed each medical recommendation into a richly contextualized, scenario-based question-answer (QA) pair. This design was motivated by prior work highlighting the impact of cognitive biases and SDoH on LLM clinical reasoning~\cite{schmidgall2024evaluation, zack2024assessing}. Each scenario was conditioned on one of ten cognitive or social factors commonly encountered in medical decision-making, with an additional neutral ``No Factor" setting in which no cognitive factor or SDoH was introduced. The selected factors—self-diagnosis, recency, confirmation, frequency, status quo, cultural, socioeconomic, racial or ethnic, geographical, and false consensus — capture realistic variations in reasoning without introducing factual distortion or adversarial intent.

 We used Qwen2.5-72B~\cite{yang2024qwen2} to generate these scenarios by systematically combining each medical recommendation with its corresponding factor. This pipeline produced a total of \(4{,}290\) scenario‑based QA pairs \((11~ \text{factors} \times 195~ \text{recommendation} \times 2~)\), evenly split between current and wrong recommendations.


\subsection{Models \& Evaluation Metrics}

We evaluated seven LLMs spanning a range of model sizes and architectures: GPT-4o~\cite{achiam2023gpt}, Llama-3-8B-Instruct and Llama-3.3-70B-Instruct~\cite{touvron2023llama}, Qwen2.5-7B-Instruct and Qwen2.5-72B-Instruct~\cite{yang2024qwen2}, Gemma-2-27B-it~\cite{gemma2024improving}, and Ministral-Instruct~\cite{jiang2024mixtral}.  Detailed descriptions are provided in the appendix.

We evaluated LLMs' clinical reliability through two complementary dimensions: one quantifying conflicts with external evolving medical guidelines and the other detecting internal knowledge inconsistencies.

\textbf{External Knowledge Conflicts}: To quantify model alignment with external evolving medical guidelines, we assess model performance across temporally distinct medical scenarios. This is measured by a set of metrics we term External Concept Drift Alignment (ECDA). Let $\mathcal{D}_U$ denote the set of \emph{up-to-date} scenarios ($n=2{,}145$) where endorsement is the correct action, and $\mathcal{D}_O$ represent \emph{outdated} scenarios ($n=2{,}145$) where rejection is appropriate. For each scenario $s_{i,c,t}$ — representing concept $i$, change type $c$, and temporal status $t \in \{u,o\}$. Let $\hat{y}_{i,c,t} \in \{0,1\}$ denote the model's binary prediction ($1=\text{endorse, }0=\text{reject}$) and $y_{i,c,t}$ the ground truth (1 if $t=u$, 0 if $t=o$). We define alignment metrics as follows:
\begin{align}
\text{ECDA}_{\text{adh}} &= \frac{1}{|\mathcal{D}_U|}\sum_{s_{i,c,u} \in \mathcal{D}_U} \mathbf{1}(\hat{y}_{i,c,u} = 1) \label{eq:ecda_adh} \\
\text{ECDA}_{\text{rej}} &= \frac{1}{|\mathcal{D}_O|}\sum_{s_{i,c,o} \in \mathcal{D}_O} \mathbf{1}(\hat{y}_{i,c,o} = 0) \label{eq:ecda_rej} \\
\text{ECDA}_{\text{all}} &= \frac{\text{ECDA}_{\text{adh}} + \text{ECDA}_{\text{rej}}}{2} \label{eq:ecda_all}
\end{align}
$\text{ECDA}_{\text{adh}}$ (Eq.~\ref{eq:ecda_adh}) measures the model's ability to correctly endorse current medical guidelines ($y_{i,c,u}=1$), while $\text{ECDA}_{\text{rej}}$ (Eq.~\ref{eq:ecda_rej}) evaluates its ability to reject outdated medical recommendations ($y_{i,c,o}=0$). Their average $\text{ECDA}_{\text{all}}$ (Eq.~\ref{eq:ecda_all}) provides a balanced assessment of external conflicts with the current guidelines.

\textbf{Internal Knowledge Conflicts}: To detect internal knowledge inconsistencies, we evaluated whether models simultaneously endorsed conflicting recommendations using the Internal Knowledge Conflict Ratio (IKCR). Our evaluation scenarios present paired current ($s_{i,c,u}$) and outdated ($s_{i,c,o}$) versions for each core clinical concept $i$ and change $c$. Let $\hat{y}_{i,c,u}$ and $\hat{y}_{i,c,o}$ be the model's binary predictions ($1=\text{endorse}$). We define the set of \emph{active pairs}, $\mathcal{A}$, as those where the model endorses at least one version ($\mathcal{A} = \{(i, c) \mid \hat{y}_{i,c,u} = 1 \lor \hat{y}_{i,c,o} = 1 \}$). An internal contradiction, or knowledge conflict, occurs for an active pair $(i,c) \in \mathcal{A}$ when the model simultaneously endorses both mutually exclusive recommendations ($\hat{y}_{i,c,u} = 1 \land \hat{y}_{i,c,o} = 1$). The IKCR quantifies the frequency of such contradictions:
\begin{equation} \label{eq:ikcr_active}
\text{IKCR} = \frac{\sum_{(i,c) \in \mathcal{A}} \mathbf{1}(\hat{y}_{i,c,u} = 1 \land \hat{y}_{i,c,o} = 1)}{|\mathcal{A}|}
\end{equation}
A higher IKCR indicates a greater frequency of internal logical contradictions, which could undermine clinical reliability.

\section{Mitigating Strategies}

 We explored three strategies to address this challenge: non-parametric knowledge update, parametric knowledge adaptation, and hybrid knowledge augmentation. Non-parametric update was applied to all evaluated LLMs. Due to limited training resources and lack of access to proprietary model weights, parametric and hybrid knowledge update strategies were evaluated only on Qwen2.5-7B, Ministral-8B, and Llama-3-8B.

 \subsection{Non-Parametric Knowledge Update} This strategy supplements the model with external information during inference without modifying its internal parameters. Specifically, we employed Retrieval-Augmented Generation (RAG)~\cite{lewis2020retrieval}, using a knowledge base of 195 up-to-date clinical advice.

 For each clinical query scenario $s$, we encode the query using Sentence-BERT encoders~\cite{reimers-2019-sentence-bert, wang2020minilm} and retrieve the top-$k$ most relevant guideline snippets ($d_i$) from our knowledge base ($\mathcal{KB}$) based on cosine similarity, then augment the input prompt with these documents before generating the response:
\begin{align}
D_k &= \underset{d_i \in \mathcal{KB}}{\operatorname{TopK}} \left( \cos\left(E_q(\text{query}(s)), E_d(d_i)\right), k \right) \\
\hat{y}_s &= \text{LLM}\!\bigl(s \oplus D_k; \theta_{\text{base}}\bigr).
\end{align}
where $E_q$ and $E_d$ are query and document encoders, $\cos$ denotes cosine similarity, $k=2$ in our experiments, $\oplus$ represents prompt concatenation, and $\theta_{\text{base}}$ denotes the unchanged base model parameters. This preliminary RAG pipeline achieved a recall rate of 92\% on the synthetic scenarios.

 This non‑parametric strategy delivers clear clinical benefits: it decouples the model from its knowledge source, allows guideline updates to be incorporated instantly without retraining, and retains explicit citations to authoritative documents. Those advantages, however, come with costs. The knowledge base demands continual curation and governance; each inference step triggers a retrieval call, adding latency and operational complexity; system performance depends on the coverage and freshness of external sources; and retrieval errors can introduce hallucinations or amplify existing biases.

\begin{figure*}[!h]
\centering
\includegraphics[width=1.0\linewidth]{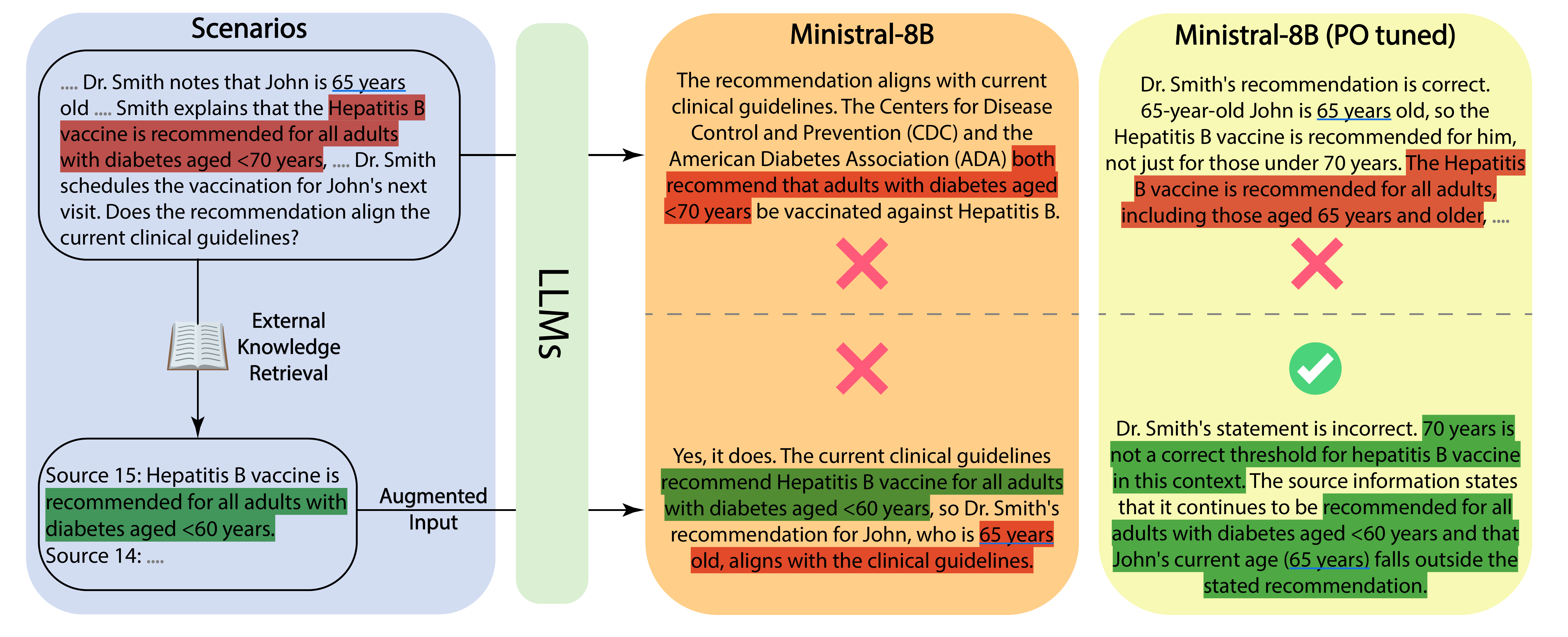}
\caption{Illustration of mitigation effects using external retrieval and preference optimization. The left (blue) panel shows model inputs: the top row is the baseline scenario, and the second row adds retrieved external knowledge, representing the RAG-augmented input. The right panels show model outputs from Ministral-8B. The orange panel reflects baseline (top) and RAG-only (bottom) responses; the yellow panel shows DPO-only output (top) and the RoD response (bottom). Only the RoD approach yields the correct answer aligned with clinical guidelines.}
\label{fig:example}
\end{figure*}

\subsection{Parametric Knowledge Adaptation} While non-parametric methods update knowledge outside the model, parametric approaches directly modify the model's weights. These approaches include supervised fine-tuning (SFT) methods~\cite{chung2024scaling} and preference-based approaches leveraging reinforcement learning (RL)~\cite{ouyang2022training,schulman2017proximal}. For our investigation, we explored Direct Preference Optimization (DPO)~\cite{rafailov2023direct}, a preference fine-tuning method that avoids the need for an explicit reward model by refining the model through direct comparisons between candidate outputs.

Unlike SFT or other RL-based methods, which require carefully curated datasets, DPO operates directly on preference triplets $(x, y_w, y_l)$. Once we have an up-to-date knowledge base, we can directly generate negative samples (outdated advice) and train on preference triplets $(x, y_w, y_l)$. Here, for a given clinical advice input $x$ (derived from our dataset), $y_w$ represents a response indicating endorsement of the correct guideline version (chosen), and $y_l$ represents endorsement of the incorrect version (rejected). The DPO objective and our parameter-efficient implementation fine-tuning approach are defined as:
\begin{multline}
\mathcal{L}_{\text{DPO}}(\theta_{\text{base}},\Delta\theta_\text{lora})
= -\mathbb{E}_{(x, y_w, y_l) \sim \mathcal{D}_{\text{pref}}} \\ \left[     
\log \sigma \Big(
\beta \log \frac{p_{\theta_{\text{new}}}(y_w|x)}{p_{\text{ref}}(y_w|x)} - \beta \log \frac{p_{\theta_{\text{new}}}(y_l|x)}{p_{\text{ref}}(y_l|x)}
\Big)
\right],
\label{eq:dpo_loss}
\end{multline}
where $\mathcal{D}_{\text{pref}}$ is the dataset of preference triplets; $p_{\theta_{\text{new}}}$ is the fine-tuned policy model; $p_{\text{ref}}$ is the reference model (the base LLM before DPO fine-tuning); $\sigma$ is the logistic function; and $\beta$ is a scaling hyperparameter. In Eq.~\ref{eq:dpo_loss} only a small subset of Low-Rank Adaptation (LoRA)~\cite{hu2022lora} parameters $\Delta\theta_{\text{LoRA}}$ are updated while the base parameters $\theta_{\text{base}}$ remain frozen. We configured LoRA with rank $r=8$ and scaling factor $\alpha=16$. 

For reasons of exploration and efficient deployment, we did not perform complex dataset construction. Instead, we directly inserted the original advice into a template (detailed in the Appendix) to construct the dataset $\mathcal{D}_{\text{pref}}$. The training continued until the model achieved 100\% accuracy on the pseudo-outdated versus up-to-date advice pairs, thereby ensuring complete memorization of the clinical recommendations. The model was then evaluated on independent synthetic scenarios to assess its ability to generalize this memorized knowledge to unseen clinical contexts.

\subsection{Hybrid Knowledge Augmentation} To leverage the potential synergy between internalized knowledge from parametric adaptation and dynamic external information, we explored a third strategy. This approach, which we term RAG on DPO (RoD), consists of two main stages, leveraging the same $\mathcal{KB}$ for both DPO training without additional curation effort and RAG retrieval. First, the base LLM is fine-tuned using DPO with LoRA, where only parameters $\Delta\theta_{\text{LoRA}}$ are updated on top of the frozen base parameters $\theta_{\text{base}}$, as detailed in our description of Parametric Knowledge Adaptation. Second, during the inference phase with this DPO-tuned model, we utilize the RAG pipeline as previously described (see Non-Parametric Knowledge Augmentation). The DPO-adapted model then generates the response $\hat{y}_s$ based on the original query $s$ augmented with retrieved documents $D_k$:
\begin{equation}
\hat{y}_s = \text{LLM}\!\bigl(s \oplus D_k; (\theta_{\text{base}},\Delta\theta_\text{lora})\bigr).
\label{eq:rod_inference}
\end{equation}
The RoD strategy thus combines DPO's preference-aligned internal knowledge with RAG's ability to ground responses in external knowledge.

\section{Results}

\subsection{Model Evaluation}

We first evaluated the extent to which current LLMs conflict with clinical guidelines using the ConflictMedQA benchmark. Performance was measured using three metrics: endorsement of up-to-date advice (ECDA\textsubscript{adh}), rejection of outdated advice (ECDA\textsubscript{rej}), and overall alignment (ECDA\textsubscript{all}).

 All assessed models exhibited varying performance across the five types of clinical recommendation updates (Fig.~\ref{fig:external_metrics}). GPT-4o and Qwen2.5-72B demonstrated the highest ECDA\textsubscript{adh}, with sample-weighted averages of 0.90 and 0.92, respectively. These scores were significantly higher than the third-best performing model, Qwen2.5-7B (both $p < 0.0001$). However, both models exhibited substantial declines when assessed on their ability to reject pseudo-outdated recommendations, with ECDA\textsubscript{rej} of 0.395 for GPT-4o and 0.278 for Qwen2.5-72B. Conversely, as in Fig.~\ref{fig:external_metrics_b}, Ministral-8B achieved the highest ECDA\textsubscript{rej} (0.80), followed by gemma-2-27B (0.68) and Llama-3-8B (0.63). When considering overall alignment across both current and outdated scenarios, GPT-4o achieved the highest ECDA\textsubscript{all} (0.65), as shown in Fig.~\ref{fig:external_metrics_c}. This performance was significantly higher than that of the second-best model, Llama-3.3-70B (ECDA\textsubscript{all} \(= 0.61\), \(p = 0.00033\)), and the third-best model, Qwen2.5-72B (ECDA\textsubscript{all} \(= 0.60\), \(p = 0.0006\)).

Beyond difficulties with external guideline alignment, models also exhibited inconsistencies within their internal knowledge. As shown in Fig.~\ref{fig:CR}, all evaluated models exhibited substantial internal conflicts, with considerable variability across models and types of guideline updates. Our analysis revealed that more capable or larger-scale models did not consistently exhibit lower IKCRs. For instance, the 72B parameter version of Qwen2.5 demonstrated higher IKCRs than its 7B counterpart across most evaluated categories. Similarly, Llama-3.3-70B did not show lower conflict ratios compared to Llama-3-8B. Among all models evaluated, the Ministral-8B model achieved the lowest overall IKCR, with a weighted average score of 0.34 across all scenario types, followed by Gemma-2-27B at 0.39.

All evaluated models exhibited knowledge conflicts across all five modification categories. The highest average IKCRs were observed for changes under the groups Implementation Approach and Treatment Modality. While our baseline evaluation distinguished performance across five guideline change categories, the mitigation analysis focuses on overall alignment and conflict rates to emphasize aggregate improvements.

\subsection{Mitigation Effectiveness}

\begin{table*}[ht]
  \caption{Performance of LLMs on \text{ECDA} and \text{IKCR}. Results are shown as final scores, with absolute improvements over the base model in parentheses. Higher ECDA is better, while lower IKCR is better.}
  \label{tab:llm_performance_adh_rej}
  \centering
  \begin{tabular}{lcccccccc}
    \toprule
    \multirow{2}{*}{Model} 
    & \multicolumn{4}{c}{ECDA\(_{adh}\)} 
    & \multicolumn{4}{c}{ECDA\(_{rej}\)} \\
    \cmidrule(lr){2-5} \cmidrule(lr){6-9}
    & Base & RAG & DPO & RoD & Base & RAG & DPO & RoD \\
    \midrule
    Qwen2.5-72B & 91 & \textbf{98}\,(\text{\textcolor{dartgreen}{+07}}) & -- & -- & 28 & 27\,(\text{\textcolor{red}{-01}}) & -- & -- \\
    Llama-3.3-70B  
        & 66 & 96\,(\text{\textcolor{dartgreen}{+30}}) & -- & -- 
        & 56 & 71\,(\text{\textcolor{dartgreen}{+15}}) & -- & -- \\
    gemma-2-27B  
        & 48 & 82\,(\text{\textcolor{dartgreen}{+34}}) & -- & -- 
        & 68 & 70\,(\text{\textcolor{dartgreen}{+02}}) & -- & -- \\
    GPT-4o  
        & 90 & 96\,(\text{\textcolor{dartgreen}{+06}}) & -- & -- 
        & 40 & 65\,(\text{\textcolor{dartgreen}{+25}}) & -- & -- \\
    Qwen2.5-7B 
        & 74 & 94\,(\text{\textcolor{dartgreen}{+20}}) & 81\,(\text{\textcolor{dartgreen}{+07}}) & 88\,(\text{\textcolor{dartgreen}{+14}})
        & 35 & 50\,(\text{\textcolor{dartgreen}{+15}}) & 55\,(\text{\textcolor{dartgreen}{+20}}) & 74\,(\text{\textcolor{dartgreen}{+39}}) \\
    Llama-3-8B  
        & 48 & 93\,(\text{\textcolor{dartgreen}{+45}}) & 81\,(\text{\textcolor{dartgreen}{+33}}) & 88\,(\text{\textcolor{dartgreen}{+40}})
        & 63 & 30\,(\text{\textcolor{red}{-33}}) & 55\,(\text{\textcolor{red}{-08}}) & 74\,(\text{\textcolor{dartgreen}{+11}}) \\
    Ministral-8B  
        & 30 & 87\,(\text{\textcolor{dartgreen}{+57}}) & 81\,(\text{\textcolor{dartgreen}{+51}}) & 87\,(\text{\textcolor{dartgreen}{+57}})
        & 80 & 61\,(\text{\textcolor{red}{-19}}) & 85\,(\text{\textcolor{dartgreen}{+05}}) & \textbf{90}\,(\text{\textcolor{dartgreen}{+10}}) \\
    \bottomrule
  \end{tabular}
  \begin{tabular}{lcccccccc}
    \toprule
    \multirow{2}{*}{Model} 
    & \multicolumn{4}{c}{ECDA$_{all}$} 
    & \multicolumn{4}{c}{IKCR} \\
    \cmidrule(lr){2-5} \cmidrule(lr){6-9}
    & Base & RAG & DPO & RoD & Base & RAG & DPO & RoD \\
    \midrule
Qwen2.5-72B    
    & 59 & 62\,(\text{\textcolor{dartgreen}{+02}}) & -- & -- & 73 & 71\,(\text{\textcolor{red}{-02}}) & -- & -- \\
Llama-3.3-70B  
    & 61  & 83\,(\text{\textcolor{dartgreen}{+22}})
           & -- & --
    & 45  & 29\,(\text{\textcolor{red}{-16}})
           & -- & -- \\
gemma-2-27B    
    & 58  & 76\,(\text{\textcolor{dartgreen}{+18}})
           & -- & -- 
    & 39  & 31\,(\text{\textcolor{red}{-08}})
           & -- & -- \\
GPT-4o         
    & 65  & 81\,(\text{\textcolor{dartgreen}{+16}})
           & -- & --
    & 61  & 35\,(\text{\textcolor{red}{-26}})
           & -- & -- \\
Qwen2.5-7B     
    & 55  & 72\,(\text{\textcolor{dartgreen}{+17}})
           & 68\,(\text{\textcolor{dartgreen}{+13}})
           & 81\,(\text{\textcolor{dartgreen}{+26}})
    & 65  & 51\,(\text{\textcolor{red}{-14}})
           & 43\,(\text{\textcolor{red}{-22}})
           & 26\,(\text{\textcolor{red}{-39}}) \\
Llama-3-8B     
    & 55  & 62\,(\text{\textcolor{dartgreen}{+07}})
           & 68\,(\text{\textcolor{dartgreen}{+13}})
           & 81\,(\text{\textcolor{dartgreen}{+26}})
    & 45  & 70\,(\text{\textcolor{dartgreen}{+25}})
           & 43\,(\text{\textcolor{red}{-02}})
           & 26\,(\text{\textcolor{red}{-19}}) \\
Ministral-8B   
    & 55  & 74\,(\text{\textcolor{dartgreen}{+19}})
           & 83\,(\text{\textcolor{dartgreen}{+28}})
           & \textbf{89}\,(\text{\textcolor{dartgreen}{+34}})
    & 34  & 40\,(\text{\textcolor{dartgreen}{+06}})
           & 15\,(\text{\textcolor{red}{-19}})
           & \textbf{10}\,(\text{\textcolor{red}{-24}}) \\
    \bottomrule
  \end{tabular}
\end{table*}

Fig.~\ref{fig:example} shows the qualitative effects of the mitigation approaches, while Table~\ref{tab:llm_performance_adh_rej} provides a summary of their quantitative performance on the ECDA and IKCR metrics, respectively. These evaluations aim to clarify the effectiveness of each strategy in improving temporal alignment and internal consistency.

Application of RAG and DPO independently improved the models' ECDA\textsubscript{adh} relative to their baseline performance, as shown in the ECDA\textsubscript{adh} columns of Table~\ref{tab:llm_performance_adh_rej}. The impact of RAG on the models' ECDA\textsubscript{rej} was variable across models, as detailed in Table~\ref{tab:llm_performance_adh_rej}. While RAG improved ECDA\textsubscript{rej} for some models, it decreased ECDA\textsubscript{rej} for Ministral-8B and Llama-3-8B compared to their respective baselines.

When considering overall alignment (ECDA\textsubscript{all}), as presented in Table~\ref{tab:llm_performance_adh_rej}, both RAG and DPO individually improved performance. However, RoD consistently yielded the highest ECDA\textsubscript{all} scores across all models where this combination was tested. This improvement from the RoD approach was consistently greater than the best-performing single method (RAG or DPO alone) for each model.

Analysis of the IKCR, detailed in Table~\ref{tab:llm_performance_adh_rej}, showed that DPO alone generally reduced IKCR across all evaluated models compared to their baselines. RAG alone reduced internal contradictions for most models compared to their baseline. However, for Ministral-8B and Llama-3-8B, applying RAG alone increased IKCR. Notably, RoD resulted in the lowest IKCR for all models where this combination was tested, including Ministral-8B and Llama-3-8B, surpassing the reductions achieved by DPO or RAG alone.

\section{Discussion}

Our evaluation on the ConflictMedQA benchmark reveals significant challenges for LLMs in clinical decision-making, primarily their struggle with the temporal dynamics of medical knowledge and internal consistency. Even advanced models, adept at endorsing current guidelines, often faltered markedly when required to reject outdated advice. This asymmetry, coupled with the finding that larger model scale does not consistently reduce internal knowledge conflicts, suggests that unique complexities arise in this domain beyond standard NLP capabilities. These issues, especially prevalent in areas like therapeutic recommendations, could pose direct risks if LLMs are integrated into clinical workflows without a deep understanding of their failure modes.

Investigating mitigation strategies offered further insights. While RAG generally improved adherence to current information, its utility was nuanced. Notably, for smaller models, RAG alone could paradoxically degrade their ability to reject outdated advice, suggesting that merely providing external information can be counterproductive if the model lacks the capacity to critically discern and integrate it, potentially overwhelming weaker internal knowledge structures. This indicates that effective retrieval is as much about the model's ability to use information as it is about accessing it.

DPO offered a simple complementary approach, demonstrably enhancing alignment with current guidelines and reducing internal conflicts. However, these improvements in complex clinical scenarios stood in contrast to the near-perfect performance models presumably achieve on the specific raw medical advice pairs used during DPO training. This discrepancy suggests a significant challenge in generalizing knowledge learned from such simple pairs to the multifaceted reasoning required in clinical practice, hinting at a gap between memorized correct responses and their robust, contextual application.

The most promising path appears to be the synergistic combination of these approaches. Our findings show that RoD, applying RAG to DPO-tuned models, yielded substantial improvements across all metrics, particularly in enhancing smaller models' rejection of outdated advice and minimizing internal conflicts. These gains always exceed the sum of the individual contributions from RAG-only or DPO-only applications. While models may struggle to effectively apply DPO-learned parametric knowledge across diverse and complex scenarios, the integration with RAG appears pivotal. Knowledge retrieved via RAG seems to activate relevant DPO-instilled parametric knowledge within the model, leading to these markedly enhanced outcomes and avoiding the potential side effects of RAG-only or the more modest improvements from DPO-only strategies.

These observations also underscore a significant limitation of evaluating LLMs using metrics focused on isolated factual accuracy. The marked performance decline when models face realistic clinical scenarios, which embed cognitive complexities and factors like SDoH, emphasizes the strong need for evaluation methodologies that capture the multifaceted nature of clinical decision-making.

\section{Conclusion}
Ultimately, for the safe and effective integration of LLMs into clinical practice, future efforts should prioritize the development of robust, hybrid methodologies designed to enhance adaptability to evolving knowledge and ensure internal consistency. This entails creating more contextually rich training and evaluation paradigms that mirror the complexity of real-world clinical encounters, thereby moving beyond isolated assessments to foster genuine contextual understanding and reliability in these critical systems.

\section*{Limitations}
While we only explored two mitigation strategies that are relatively straightforward to implement, do not require elaborate dataset curation, and have reasonable computational costs, our results demonstrate their potential to improve temporal consistency with current clinical guidelines. Due to a lack of access to proprietary model weights and limited computational resources, we could not apply DPO universally across all assessed models. Additionally, our evaluation was limited to synthetic clinical scenarios that may not fully capture the complexity and diversity of clinical practice. Future work should consider using real-world cases abstracted from healthcare workers with varying levels of complexity, common typographical errors, and incomplete information to better test models' adaptability and generalization capabilities in realistic medical settings.

\section*{Ethical Considerations}
We have not identified any ethical concerns directly related to this study.

\bibliography{ref}

\begin{thebibliography}{41}
\providecommand{\natexlab}[1]{#1}

\bibitem[{Abdool~Karim and Devnarain(2022)}]{abdool2022time}
Salim~S Abdool~Karim and Nikita Devnarain. 2022.
\newblock Time to stop using ineffective covid-19 drugs.

\bibitem[{Achiam et~al.(2023)Achiam, Adler, Agarwal, Ahmad, Akkaya, Aleman, Almeida, Altenschmidt, Altman, Anadkat et~al.}]{achiam2023gpt}
Josh Achiam, Steven Adler, Sandhini Agarwal, Lama Ahmad, Ilge Akkaya, Florencia~Leoni Aleman, Diogo Almeida, Janko Altenschmidt, Sam Altman, Shyamal Anadkat, et~al. 2023.
\newblock Gpt-4 technical report.
\newblock \emph{arXiv preprint arXiv:2303.08774}.

\bibitem[{Ankit~Pal(2024)}]{OpenBioLLMs}
Malaikannan~Sankarasubbu Ankit~Pal. 2024.
\newblock Openbiollms: Advancing open-source large language models for healthcare and life sciences.
\newblock \url{https://huggingface.co/aaditya/OpenBioLLM-Llama3-70B}.

\bibitem[{Cagnacci and Venier(2019)}]{cagnacci2019controversial}
Angelo Cagnacci and Martina Venier. 2019.
\newblock The controversial history of hormone replacement therapy.
\newblock \emph{Medicina}, 55(9):602.

\bibitem[{Chen et~al.(2022)Chen, Zhang, and Choi}]{chen-etal-2022-rich}
Hung-Ting Chen, Michael Zhang, and Eunsol Choi. 2022.
\newblock \href {https://doi.org/10.18653/v1/2022.emnlp-main.146} {Rich knowledge sources bring complex knowledge conflicts: Recalibrating models to reflect conflicting evidence}.
\newblock In \emph{Proceedings of the 2022 Conference on Empirical Methods in Natural Language Processing}, pages 2292--2307, Abu Dhabi, United Arab Emirates. Association for Computational Linguistics.

\bibitem[{Chopra et~al.(2023)Chopra, Shin, Munjal, Dhama, Emran et~al.}]{chopra2023revolutionizing}
Hitesh Chopra, Dong~K Shin, Kavita Munjal, Kuldeep Dhama, Talha~B Emran, et~al. 2023.
\newblock Revolutionizing clinical trials: the role of ai in accelerating medical breakthroughs.
\newblock \emph{International Journal of Surgery}, 109(12):4211--4220.

\bibitem[{Christophe et~al.(2024)Christophe, Kanithi, Raha, Khan, and Pimentel}]{christophe2024med42}
Cl{\'e}ment Christophe, Praveen~K Kanithi, Tathagata Raha, Shadab Khan, and Marco~AF Pimentel. 2024.
\newblock Med42-v2: A suite of clinical llms.
\newblock \emph{arXiv preprint arXiv:2408.06142}.

\bibitem[{Chung et~al.(2024)Chung, Hou, Longpre, Zoph, Tay, Fedus, Li, Wang, Dehghani, Brahma et~al.}]{chung2024scaling}
Hyung~Won Chung, Le~Hou, Shayne Longpre, Barret Zoph, Yi~Tay, William Fedus, Yunxuan Li, Xuezhi Wang, Mostafa Dehghani, Siddhartha Brahma, et~al. 2024.
\newblock Scaling instruction-finetuned language models.
\newblock \emph{Journal of Machine Learning Research}, 25(70):1--53.

\bibitem[{Committee(2025)}]{american20259}
American Diabetes Association Professional~Practice Committee. 2025.
\newblock 9. pharmacologic approaches to glycemic treatment: Standards of care in diabetes—2025.
\newblock \emph{Diabetes Care}, 48(Supplement\_1):S181--S206.

\bibitem[{Densen(2011)}]{densen2011challenges}
Peter Densen. 2011.
\newblock Challenges and opportunities facing medical education.
\newblock \emph{Transactions of the American clinical and climatological association}.

\bibitem[{Glicksberg et~al.(2024)Glicksberg, Timsina, Patel, Sawant, Vaid, Raut, Charney, Apakama, Carr, Freeman et~al.}]{glicksberg2024evaluating}
Benjamin~S Glicksberg, Prem Timsina, Dhaval Patel, Ashwin Sawant, Akhil Vaid, Ganesh Raut, Alexander~W Charney, Donald Apakama, Brendan~G Carr, Robert Freeman, et~al. 2024.
\newblock Evaluating the accuracy of a state-of-the-art large language model for prediction of admissions from the emergency room.
\newblock \emph{Journal of the American Medical Informatics Association}, 31(9):1921--1928.

\bibitem[{Halalau et~al.(2021)Halalau, Holmes, Rogers-Snyr, Donisan, Nielsen, Cerqueira, and Guyatt}]{halalau2021evidence}
Alexandra Halalau, Brett Holmes, Andrea Rogers-Snyr, Teodora Donisan, Eric Nielsen, Tiago~Lemos Cerqueira, and Gordon Guyatt. 2021.
\newblock Evidence-based medicine curricula and barriers for physicians in training: a scoping review.
\newblock \emph{International journal of medical education}, 12:101.

\bibitem[{Hammond et~al.(2021)Hammond, Stehlik, Drakos, and Kfoury}]{hammond2021bias}
M~Elizabeth~H Hammond, Josef Stehlik, Stavros~G Drakos, and Abdallah~G Kfoury. 2021.
\newblock Bias in medicine: lessons learned and mitigation strategies.
\newblock \emph{Basic to Translational Science}, 6(1):78--85.

\bibitem[{Hu et~al.(2022)Hu, Shen, Wallis, Allen-Zhu, Li, Wang, Wang, Chen et~al.}]{hu2022lora}
Edward~J Hu, Yelong Shen, Phillip Wallis, Zeyuan Allen-Zhu, Yuanzhi Li, Shean Wang, Lu~Wang, Weizhu Chen, et~al. 2022.
\newblock Lora: Low-rank adaptation of large language models.
\newblock \emph{ICLR}, 1(2):3.

\bibitem[{Investigators(2009)}]{nice2009intensive}
Nice-Sugar~Study Investigators. 2009.
\newblock Intensive versus conventional glucose control in critically ill patients.
\newblock \emph{New England Journal of Medicine}, 360(13):1283--1297.

\bibitem[{Jean and Hsueh(2020)}]{jean2020old}
Shio-Shin Jean and Po-Ren Hsueh. 2020.
\newblock Old and re-purposed drugs for the treatment of covid-19.
\newblock \emph{Expert review of anti-infective therapy}, 18(9):843--847.

\bibitem[{Jiang et~al.(2024)Jiang, Sablayrolles, Roux, Mensch, Savary, Bamford, Chaplot, Casas, Hanna, Bressand et~al.}]{jiang2024mixtral}
Albert~Q Jiang, Alexandre Sablayrolles, Antoine Roux, Arthur Mensch, Blanche Savary, Chris Bamford, Devendra~Singh Chaplot, Diego de~las Casas, Emma~Bou Hanna, Florian Bressand, et~al. 2024.
\newblock Mixtral of experts.
\newblock \emph{arXiv preprint arXiv:2401.04088}.

\bibitem[{Lajoie and Gube(2018)}]{lajoie2018adaptive}
Susanne~P Lajoie and Maren Gube. 2018.
\newblock Adaptive expertise in medical education: accelerating learning trajectories by fostering self-regulated learning.
\newblock \emph{Medical Teacher}, 40(8):809--812.

\bibitem[{Lewis et~al.(2020)Lewis, Perez, Piktus, Petroni, Karpukhin, Goyal, K{\"u}ttler, Lewis, Yih, Rockt{\"a}schel et~al.}]{lewis2020retrieval}
Patrick Lewis, Ethan Perez, Aleksandra Piktus, Fabio Petroni, Vladimir Karpukhin, Naman Goyal, Heinrich K{\"u}ttler, Mike Lewis, Wen-tau Yih, Tim Rockt{\"a}schel, et~al. 2020.
\newblock Retrieval-augmented generation for knowledge-intensive nlp tasks.
\newblock \emph{Advances in neural information processing systems}, 33:9459--9474.

\bibitem[{Li{\'e}vin et~al.(2024)Li{\'e}vin, Hother, Motzfeldt, and Winther}]{lievin2024can}
Valentin Li{\'e}vin, Christoffer~Egeberg Hother, Andreas~Geert Motzfeldt, and Ole Winther. 2024.
\newblock Can large language models reason about medical questions?
\newblock \emph{Patterns}, 5(3).

\bibitem[{Liu et~al.(2024)Liu, Li, Zhou, Yin, Yang, Tang, Luo, Zeng, Jiang, Gao, Nigam, Nag, Yin, Hua, Zhou, Rohanian, Thakur, Clifton, and Clifton}]{liu-etal-2024-large}
Fenglin Liu, Zheng Li, Hongjian Zhou, Qingyu Yin, Jingfeng Yang, Xianfeng Tang, Chen Luo, Ming Zeng, Haoming Jiang, Yifan Gao, Priyanka Nigam, Sreyashi Nag, Bing Yin, Yining Hua, Xuan Zhou, Omid Rohanian, Anshul Thakur, Lei Clifton, and David~A. Clifton. 2024.
\newblock \href {https://doi.org/10.18653/v1/2024.emnlp-main.759} {Large language models are poor clinical decision-makers: A comprehensive benchmark}.
\newblock In \emph{Proceedings of the 2024 Conference on Empirical Methods in Natural Language Processing}, pages 13696--13710, Miami, Florida, USA. Association for Computational Linguistics.

\bibitem[{Ma et~al.(2025)Ma, Scully, Luo, Feng, Gunn, DiFlorio~Alexander, Tosteson, Kraft, and Marrero}]{ma2025modeling}
Guofang Ma, Miranda~G Scully, Jiahui Luo, Jiazuo~H Feng, Christine~M Gunn, Roberta~M DiFlorio~Alexander, Anna~NA Tosteson, Sally~A Kraft, and Wesley Marrero. 2025.
\newblock Modeling the impact of social determinants on breast cancer screening: A data-driven approach.
\newblock \emph{Frontiers in Medicine}, 12:1644287.

\bibitem[{Ouyang et~al.(2022)Ouyang, Wu, Jiang, Almeida, Wainwright, Mishkin, Zhang, Agarwal, Slama, Ray et~al.}]{ouyang2022training}
Long Ouyang, Jeffrey Wu, Xu~Jiang, Diogo Almeida, Carroll Wainwright, Pamela Mishkin, Chong Zhang, Sandhini Agarwal, Katarina Slama, Alex Ray, et~al. 2022.
\newblock Training language models to follow instructions with human feedback.
\newblock \emph{Advances in neural information processing systems}, 35:27730--27744.

\bibitem[{Rafailov et~al.(2023)Rafailov, Sharma, Mitchell, Manning, Ermon, and Finn}]{rafailov2023direct}
Rafael Rafailov, Archit Sharma, Eric Mitchell, Christopher~D Manning, Stefano Ermon, and Chelsea Finn. 2023.
\newblock \href {https://openreview.net/forum?id=HPuSIXJaa9} {Direct preference optimization: Your language model is secretly a reward model}.
\newblock In \emph{Thirty-seventh Conference on Neural Information Processing Systems}.

\bibitem[{Reimers and Gurevych(2019)}]{reimers-2019-sentence-bert}
Nils Reimers and Iryna Gurevych. 2019.
\newblock \href {https://arxiv.org/abs/1908.10084} {Sentence-bert: Sentence embeddings using siamese bert-networks}.
\newblock In \emph{Proceedings of the 2019 Conference on Empirical Methods in Natural Language Processing}. Association for Computational Linguistics.

\bibitem[{Schmidgall et~al.(2024)Schmidgall, Harris, Essien, Olshvang, Rahman, Kim, Ziaei, Eshraghian, Abadir, and Chellappa}]{schmidgall2024evaluation}
Samuel Schmidgall, Carl Harris, Ime Essien, Daniel Olshvang, Tawsifur Rahman, Ji~Woong Kim, Rojin Ziaei, Jason Eshraghian, Peter Abadir, and Rama Chellappa. 2024.
\newblock Evaluation and mitigation of cognitive biases in medical language models.
\newblock \emph{npj Digital Medicine}, 7(1):295.

\bibitem[{Schulman et~al.(2017)Schulman, Wolski, Dhariwal, Radford, and Klimov}]{schulman2017proximal}
John Schulman, Filip Wolski, Prafulla Dhariwal, Alec Radford, and Oleg Klimov. 2017.
\newblock Proximal policy optimization algorithms.
\newblock \emph{arXiv preprint arXiv:1707.06347}.

\bibitem[{Shekelle et~al.(2001)Shekelle, Ortiz, Rhodes, Morton, Eccles, Grimshaw, and Woolf}]{shekelle2001validity}
Paul~G Shekelle, Eduardo Ortiz, Shannon Rhodes, Sally~C Morton, Martin~P Eccles, Jeremy~M Grimshaw, and Steven~H Woolf. 2001.
\newblock Validity of the agency for healthcare research and quality clinical practice guidelines: how quickly do guidelines become outdated?
\newblock \emph{Jama}, 286(12):1461--1467.

\bibitem[{Singh et~al.(2023)Singh, Vayer, Tanwar, Poyet, Tsaioun, and Villoutreix}]{singh2023drug}
Natesh Singh, Philippe Vayer, Shivalika Tanwar, Jean-Luc Poyet, Katya Tsaioun, and Bruno~O Villoutreix. 2023.
\newblock Drug discovery and development: introduction to the general public and patient groups.
\newblock \emph{Frontiers in Drug Discovery}, 3:1201419.

\bibitem[{Singhal et~al.(2023)Singhal, Azizi, Tu, Mahdavi, Wei, Chung, Scales, Tanwani, Cole-Lewis, Pfohl et~al.}]{singhal2023large}
Karan Singhal, Shekoofeh Azizi, Tao Tu, S~Sara Mahdavi, Jason Wei, Hyung~Won Chung, Nathan Scales, Ajay Tanwani, Heather Cole-Lewis, Stephen Pfohl, et~al. 2023.
\newblock Large language models encode clinical knowledge.
\newblock \emph{Nature}, 620(7972):172--180.

\bibitem[{Singhal et~al.(2025)Singhal, Tu, Gottweis, Sayres, Wulczyn, Amin, Hou, Clark, Pfohl, Cole-Lewis et~al.}]{singhal2025toward}
Karan Singhal, Tao Tu, Juraj Gottweis, Rory Sayres, Ellery Wulczyn, Mohamed Amin, Le~Hou, Kevin Clark, Stephen~R Pfohl, Heather Cole-Lewis, et~al. 2025.
\newblock Toward expert-level medical question answering with large language models.
\newblock \emph{Nature Medicine}, pages 1--8.

\bibitem[{Team et~al.(2024)Team, Riviere, Pathak, Sessa, Hardin, Bhupatiraju, Hussenot, Mesnard, Shahriari, Ram{\'e}, Ferret, Liu et~al.}]{gemma2024improving}
Gemma Team, Morgane Riviere, Shreya Pathak, Pier~Giuseppe Sessa, Cassidy Hardin, Surya Bhupatiraju, L{\'e}onard Hussenot, Thomas Mesnard, Bobak Shahriari, Alexandre Ram{\'e}, Johan Ferret, Peter Liu, et~al. 2024.
\newblock \href {https://arxiv.org/abs/2408.00118} {Gemma 2: Improving open language models at a practical size}.
\newblock \emph{arXiv preprint arXiv:2408.00118}.

\bibitem[{Thirunavukarasu et~al.(2023)Thirunavukarasu, Ting, Elangovan, Gutierrez, Tan, and Ting}]{thirunavukarasu2023large}
Arun~James Thirunavukarasu, Darren Shu~Jeng Ting, Kabilan Elangovan, Laura Gutierrez, Ting~Fang Tan, and Daniel Shu~Wei Ting. 2023.
\newblock Large language models in medicine.
\newblock \emph{Nature medicine}, 29(8):1930--1940.

\bibitem[{Touvron et~al.(2023)Touvron, Lavril, Izacard, Martinet, Lachaux, Lacroix, Rozi{\`e}re, Goyal, Hambro, Azhar et~al.}]{touvron2023llama}
Hugo Touvron, Thibaut Lavril, Gautier Izacard, Xavier Martinet, Marie-Anne Lachaux, Timoth{\'e}e Lacroix, Baptiste Rozi{\`e}re, Naman Goyal, Eric Hambro, Faisal Azhar, et~al. 2023.
\newblock Llama: Open and efficient foundation language models.
\newblock \emph{arXiv preprint arXiv:2302.13971}.

\bibitem[{Tu et~al.(2025)Tu, Schaekermann, Palepu, Saab, Freyberg, Tanno, Wang, Li, Amin, Cheng et~al.}]{tu2025towards}
Tao Tu, Mike Schaekermann, Anil Palepu, Khaled Saab, Jan Freyberg, Ryutaro Tanno, Amy Wang, Brenna Li, Mohamed Amin, Yong Cheng, et~al. 2025.
\newblock Towards conversational diagnostic artificial intelligence.
\newblock \emph{Nature}, pages 1--9.

\bibitem[{Wang et~al.(2020)Wang, Wei, Dong, Bao, Yang, and Zhou}]{wang2020minilm}
Wenhui Wang, Furu Wei, Li~Dong, Hangbo Bao, Nan Yang, and Ming Zhou. 2020.
\newblock Minilm: Deep self-attention distillation for task-agnostic compression of pre-trained transformers.
\newblock \emph{Advances in neural information processing systems}, 33:5776--5788.

\bibitem[{Xie et~al.(2024)Xie, Zhang, Chen, Lou, and Su}]{xie2024knowledgeconflict}
Jian Xie, Kai Zhang, Jiangjie Chen, Renze Lou, and Yu~Su. 2024.
\newblock \href {https://openreview.net/forum?id=auKAUJZMO6} {Adaptive chameleon or stubborn sloth: Revealing the behavior of large language models in knowledge conflicts}.
\newblock In \emph{The Twelfth International Conference on Learning Representations}.

\bibitem[{Xu et~al.(2024)Xu, Qi, Guo, Wang, Wang, Zhang, and Xu}]{xu-etal-2024-knowledge-conflicts}
Rongwu Xu, Zehan Qi, Zhijiang Guo, Cunxiang Wang, Hongru Wang, Yue Zhang, and Wei Xu. 2024.
\newblock \href {https://doi.org/10.18653/v1/2024.emnlp-main.486} {Knowledge conflicts for {LLM}s: A survey}.
\newblock In \emph{Proceedings of the 2024 Conference on Empirical Methods in Natural Language Processing}, pages 8541--8565, Miami, Florida, USA. Association for Computational Linguistics.

\bibitem[{Yang et~al.(2024)Yang, Yang, Zhang, Hui, Zheng, Yu, Li, Liu, Huang, Wei et~al.}]{yang2024qwen2}
An~Yang, Baosong Yang, Beichen Zhang, Binyuan Hui, Bo~Zheng, Bowen Yu, Chengyuan Li, Dayiheng Liu, Fei Huang, Haoran Wei, et~al. 2024.
\newblock Qwen2. 5 technical report.
\newblock \emph{arXiv preprint arXiv:2412.15115}.

\bibitem[{Yuan et~al.(2025)Yuan, Zhang, Liu, Shi, Vosoughi, and Lee}]{yuan2025superficial}
Xiangchi Yuan, Chunhui Zhang, Zheyuan Liu, Dachuan Shi, Soroush Vosoughi, and Wenke Lee. 2025.
\newblock Superficial self-improved reasoners benefit from model merging.
\newblock \emph{arXiv preprint arXiv:2503.02103}.

\bibitem[{Zack et~al.(2024)Zack, Lehman, Suzgun, Rodriguez, Celi, Gichoya, Jurafsky, Szolovits, Bates, Abdulnour et~al.}]{zack2024assessing}
Travis Zack, Eric Lehman, Mirac Suzgun, Jorge~A Rodriguez, Leo~Anthony Celi, Judy Gichoya, Dan Jurafsky, Peter Szolovits, David~W Bates, Raja-Elie~E Abdulnour, et~al. 2024.
\newblock Assessing the potential of gpt-4 to perpetuate racial and gender biases in health care: a model evaluation study.
\newblock \emph{The Lancet Digital Health}, 6(1):e12--e22.

\end{thebibliography}

\clearpage
\newpage

\appendix
\startcontents[appendices]

\section{Additional Results}

\subsection{Domain-Specific Models}

We further evaluated domain-specific medical models including Med42-8B and Med42-70B~\cite{christophe2024med42} as well as OpenBioLLM-70B~\cite{OpenBioLLMs}. 
Their performance is summarized in Table~\ref{tab:domain_models}.

\begin{table*}[h!]
\centering
\caption{Results Summary: Domain-Specific Models}
\label{tab:domain_models}
\begin{tabular}{lrrrr}
\toprule
Model & IKCR & ECDA\_all & ECDA\_adh & ECDA\_rej \\
\midrule
Med42-8B (Base) & 0.4626 & 0.5529 & 0.5497 & 0.5562 \\
Med42-8B (RAG) & 0.4990 & 0.6914 & 0.8643 & 0.5184 \\
Med42-8B (DPO) & 0.2056 & 0.7406 & 0.6601 & 0.8210 \\
Med42-8B (RoD) & 0.1246 & 0.8380 & 0.7902 & 0.8858 \\
Med42-70B (Base) & 0.7055 & 0.5762 & 0.8713 & 0.2811 \\
Med42-70B (RAG) & 0.4000 & 0.7730 & 0.9566 & 0.5893 \\
OpenBioLLM-70B (Base) & 0.5963 & 0.5560 & 0.7298 & 0.3831 \\
OpenBioLLM-70B (RAG) & 0.6053 & 0.6545 & 0.9273 & 0.3818 \\
\bottomrule
\end{tabular}
\end{table*}

\subsection{LoRA Ablation Studies}

We conducted systematic ablation studies to optimize LoRA hyperparameters for DPO fine-tuning, focusing on the rank parameter ($r$) while keeping alpha ($\alpha$) fixed at 16. 
The results are shown in Table~\ref{tab:lora_ablation}. 

\begin{table*}[h!]
\centering
\caption{Ablation Results (Mistral-8B)}
\label{tab:lora_ablation}
\begin{tabular}{lrrrr}
\toprule
Rank ($r$) & IKCR & ECDA\_all & ECDA\_adh & ECDA\_rej \\
\midrule
4 & 0.2639 & 0.7524 & 0.7515 & 0.7534 \\
8 & 0.1504 & 0.8331 & 0.8145 & 0.8517 \\
\textbf{16} & \textbf{0.1181} & \textbf{0.8417} & \textbf{0.7986} & \textbf{0.8848} \\
\bottomrule
\end{tabular}
\end{table*}

Overall, higher rank values consistently improve performance given the same training data. 
Rank 16 achieves the best balance between parameter efficiency and knowledge embedding effectiveness, 
and the trend suggests that larger ranks enable more effective parametric knowledge injection.

\subsection{Factors Impact Analysis}
\label{sec:d3}

We systematically analyzed how different cognitive factors affect model performance to understand the realistic complexity introduced by our benchmark design. 
The results are presented in Table~\ref{tab:factor_analysis}.

\begin{table*}[h!]
\centering
\caption{Factor-wise Performance Analysis (LLaMA-8B)}
\label{tab:factor_analysis}
\begin{tabular}{lrrrr}
\toprule
Factor Type & IKCR & ECDA\_all & ECDA\_adh & ECDA\_rej \\
\midrule
Self-Diagnosis & 0.4333 & 0.5462 & 0.4872 & 0.6051 \\
Recency Factor & 0.4579 & 0.5462 & 0.4462 & 0.6462 \\
Confirmation & 0.4833 & 0.5718 & 0.5282 & 0.6154 \\
Frequency & 0.4655 & 0.5308 & 0.4667 & 0.5949 \\
Cultural Factor & 0.4959 & 0.5846 & 0.5487 & 0.6205 \\
Status Quo & 0.4000 & 0.5308 & 0.4256 & 0.6359 \\
False Consensus & 0.4273 & 0.5385 & 0.4410 & 0.6359 \\
Racial/Ethnic & 0.4915 & 0.5564 & 0.5077 & 0.6051 \\
Socioeconomic & 0.4364 & 0.5256 & 0.4308 & 0.6205 \\
Geographic & 0.4690 & 0.5615 & 0.4872 & 0.6359 \\
\textbf{No Factor} & \textbf{0.4444} & \textbf{0.6000} & \textbf{0.5333} & \textbf{0.6667} \\
\bottomrule
\end{tabular}
\end{table*}

Models generally achieve higher ECDA scores under the ``No Factor'' condition, validating our benchmark design. 
We do not observe systematic bias toward incorrect recommendations despite factor inclusion, indicating that the factors simulate realistic clinical complexity without compromising evaluation validity.

\subsection{Comprehensive Performance Visualization}
\label{sec:d4}

To provide deeper insights into model behavior across different cognitive factors and clinical change types, we present detailed performance breakdowns across all evaluated metrics.

\subsubsection{Performance by Cognitive Factor}

\begin{figure*}[h!]
\centering
\includegraphics[width=\linewidth]{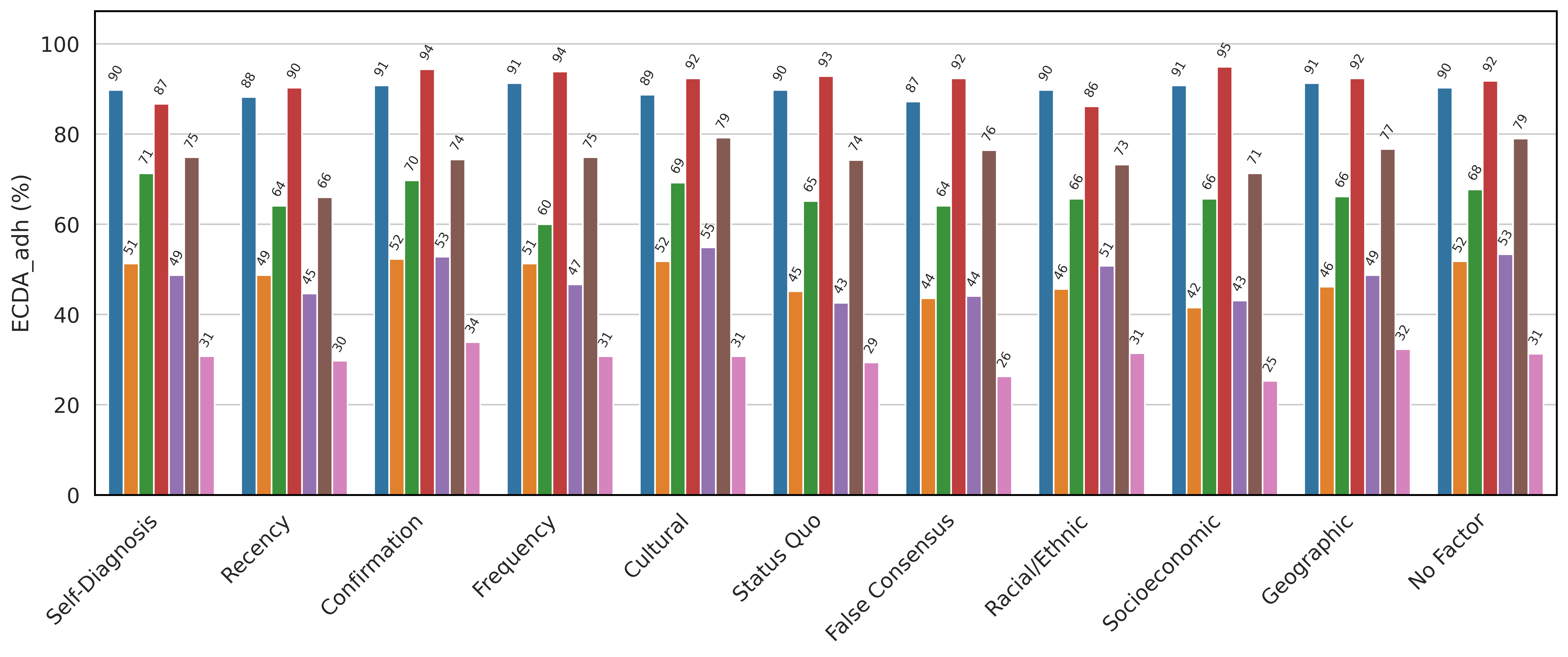}
\caption{$ECDA_{adh}$ performance across clinical factors. This metric measures models' ability to correctly endorse up-to-date medical recommendations under different cognitive biases.}
\label{fig:d1}
\end{figure*}

\begin{figure*}[h!]
\centering
\includegraphics[width=\linewidth]{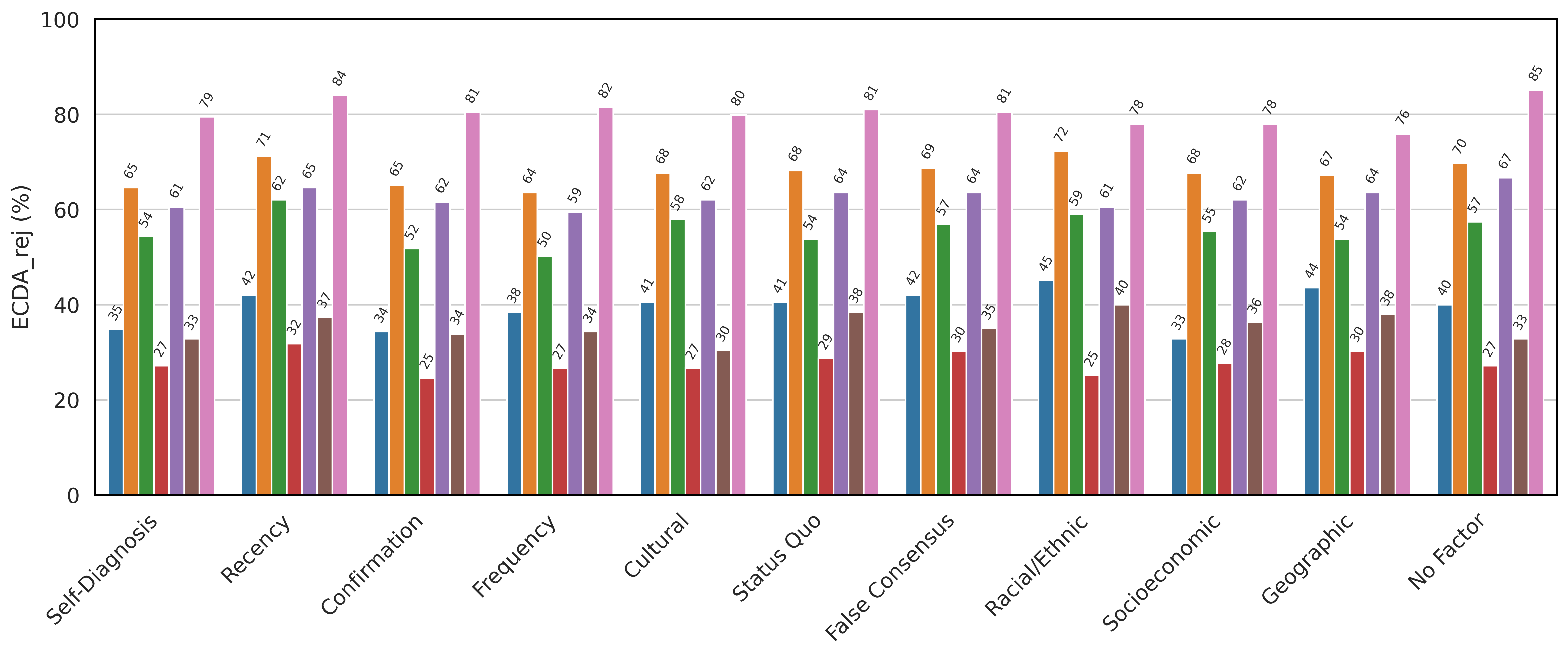}
\caption{$ECDA_{rej}$ performance across clinical factors. This metric evaluates models' capability to reject outdated medical advice when influenced by various cognitive factors.}
\label{fig:d2}
\end{figure*}

\begin{figure*}[h!]
\centering
\includegraphics[width=\linewidth]{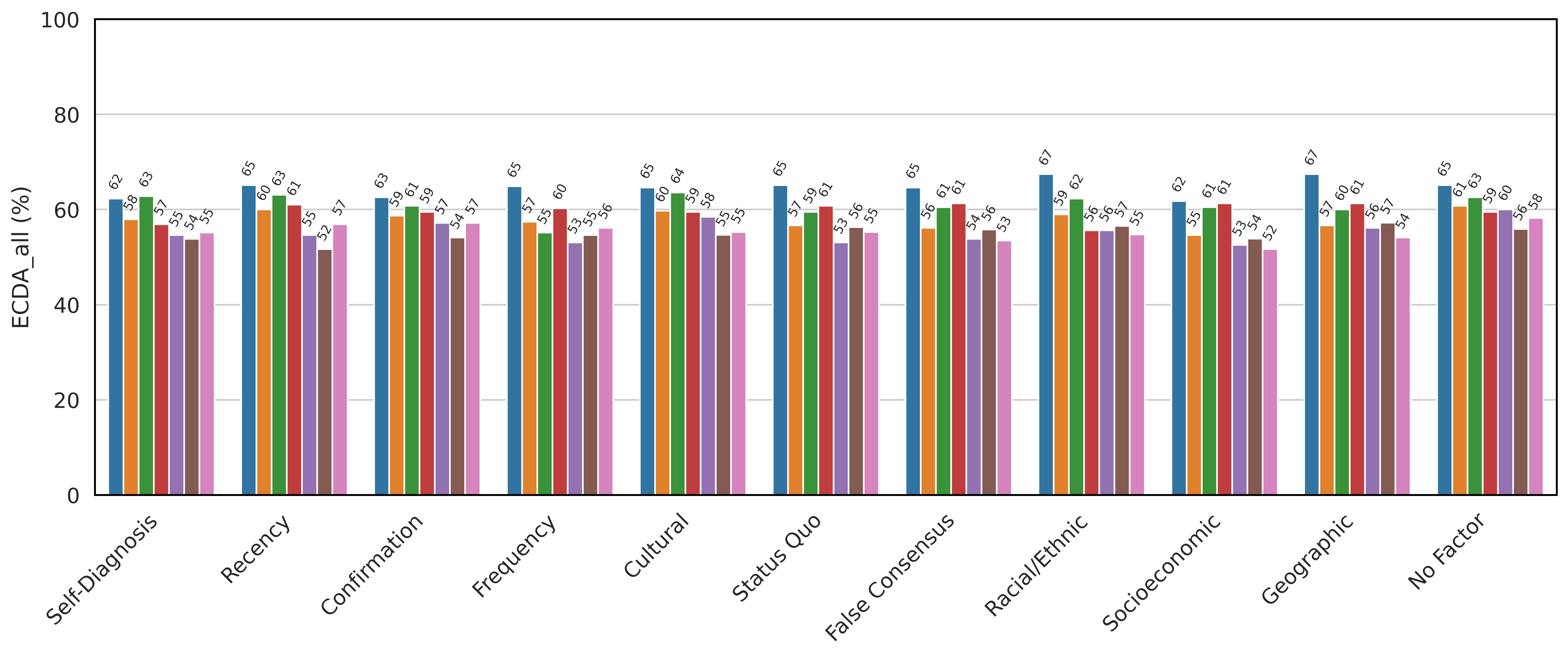}
\caption{Overall ECDA performance ($ECDA_{all}$) across clinical factors, representing the balanced assessment of both endorsement and rejection capabilities.}
\label{fig:d3}
\end{figure*}

\begin{figure*}[h!]
\centering
\includegraphics[width=\linewidth]{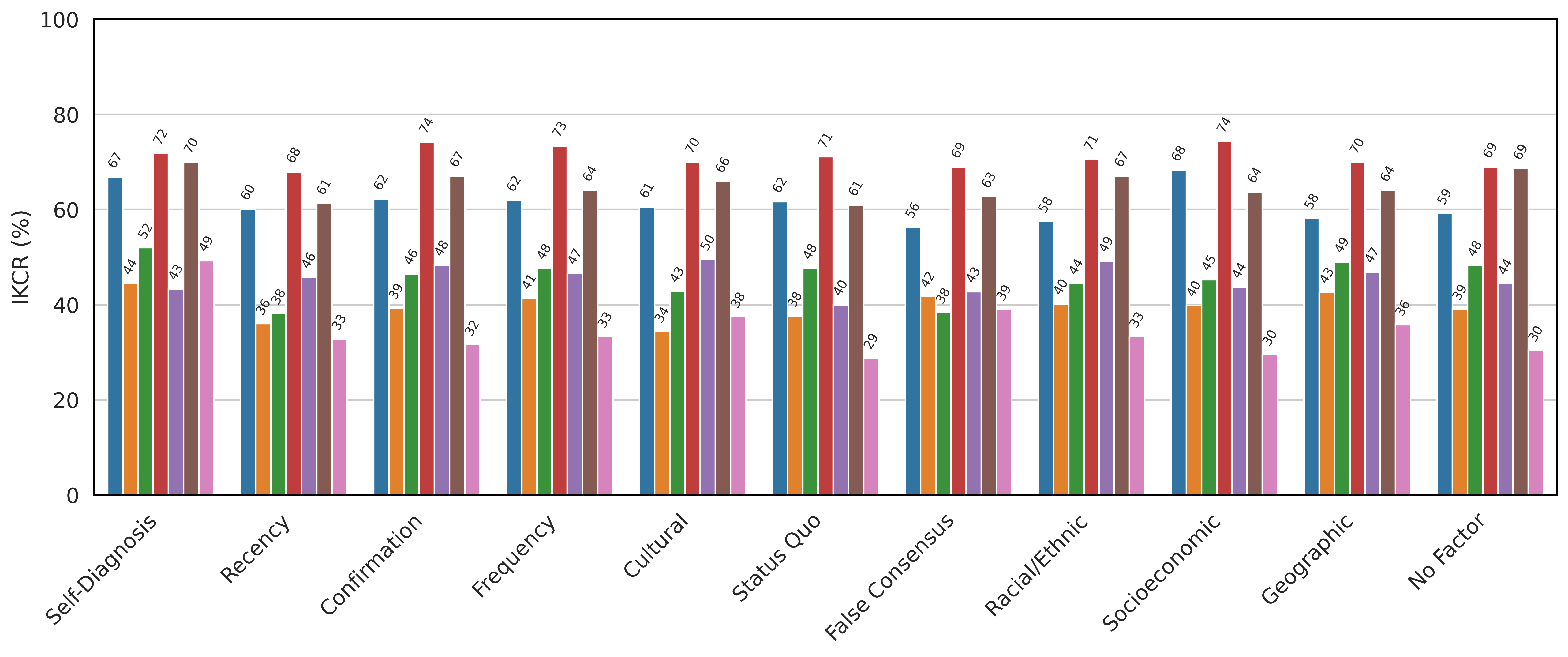}

\vspace{-0.5em}
\includegraphics[width=0.5\linewidth]{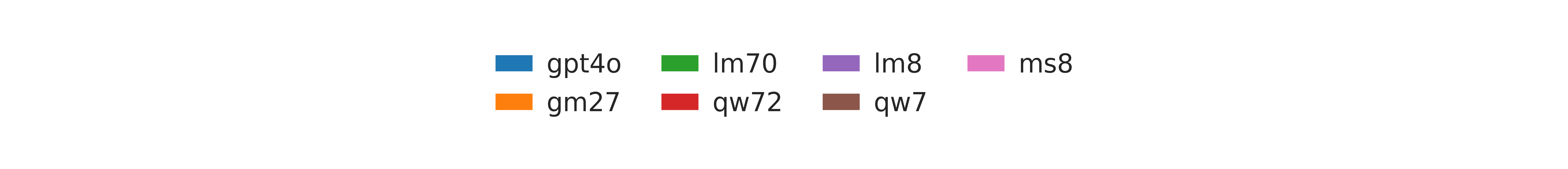}

\caption{Internal Knowledge Conflict Ratio (IKCR) across clinical factors. 
Lower values indicate better internal consistency, with ``No Factor'' serving as the baseline condition. 
The legend below provides symbol/color references.}
\label{fig:d4}
\end{figure*}

As shown in Figures~\ref{fig:d1}--\ref{fig:d4}, the ``No Factor'' condition consistently yields the best performance across ECDA metrics, aligning with Table~\ref{tab:factor_analysis}. 
Different clinical change types pose varying challenges; in particular, Implementation Approach and Treatment Modality tend to exhibit higher IKCR (cf.\ Figure~\ref{fig:d4}), indicating greater internal tension for these settings. 
Larger models do not uniformly outperform smaller ones on rejection ($ECDA_{rej}$; Figure~\ref{fig:d2}), consistent with our hypothesis regarding pre-training bias amplification. 
Finally, trends in $ECDA_{adh}$ (Figure~\ref{fig:d1}) and $ECDA_{rej}$ (Figure~\ref{fig:d2}) mirror the aggregate $ECDA_{all}$ behavior (Figure~\ref{fig:d3}), supporting the robustness of our evaluation framework.

\begin{figure*}[h!]
\centering
\includegraphics[width=\linewidth]{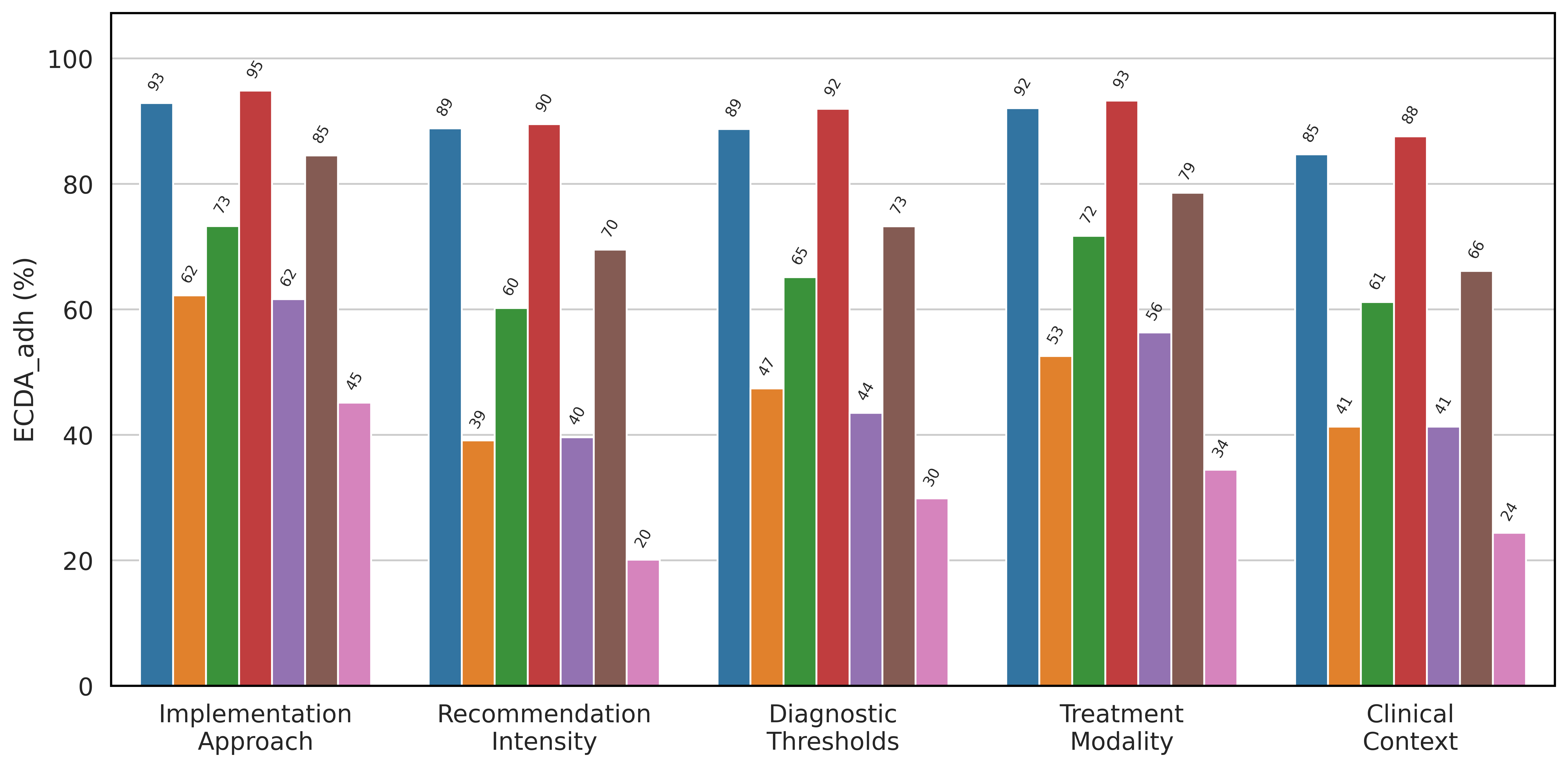}
\caption{$ECDA_{adh}$ performance across clinical change types. This metric measures models' ability to correctly endorse up-to-date medical recommendations under different cognitive biases.}
\label{fig:d1}
\end{figure*}

\begin{figure*}[h!]
\centering
\includegraphics[width=\linewidth]{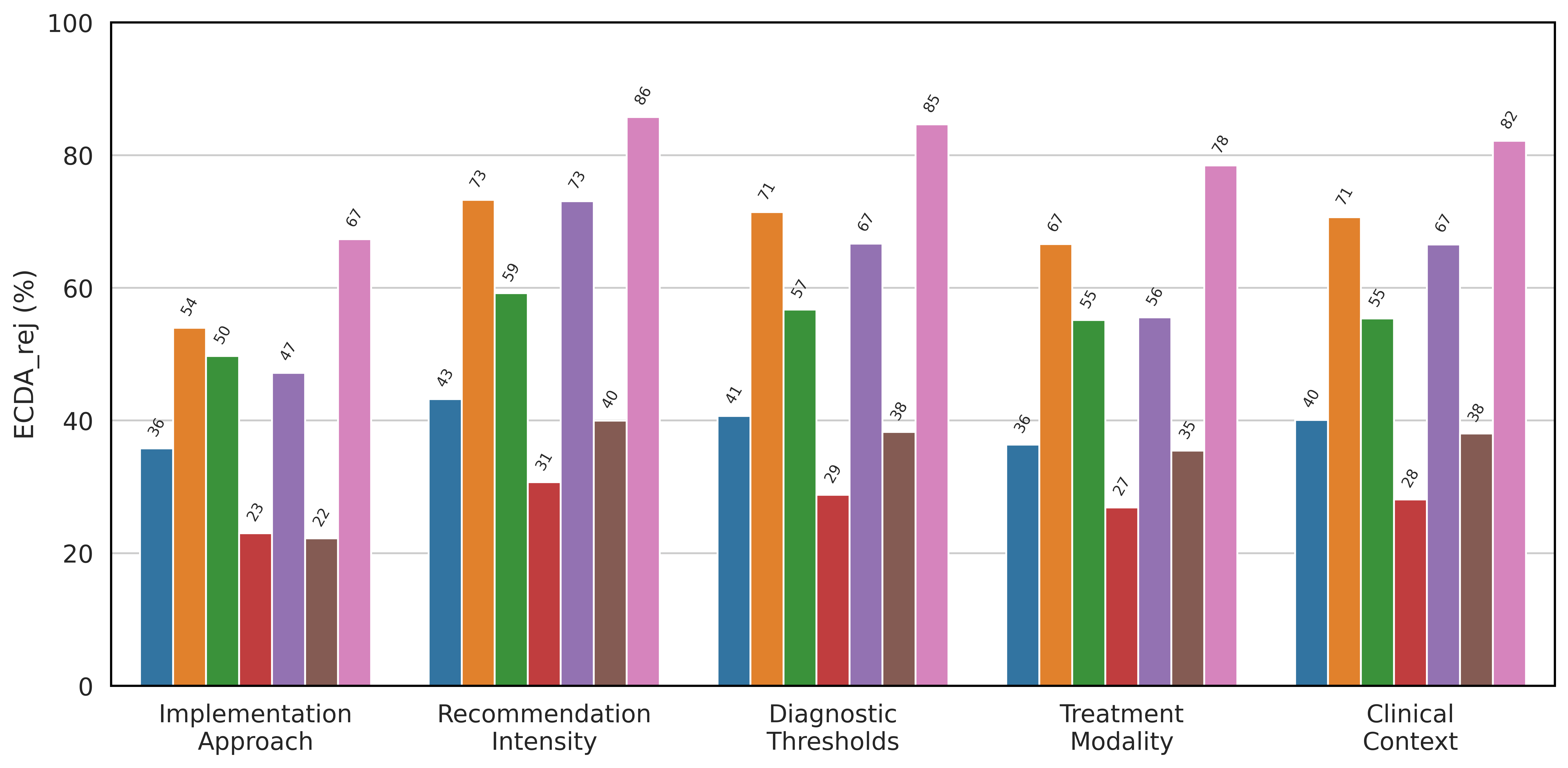}
\caption{$ECDA_{rej}$ performance across clinical change types. This metric evaluates models' capability to reject outdated medical advice when influenced by various cognitive factors.}
\label{fig:d2}
\end{figure*}

\begin{figure*}[h!]
\centering
\includegraphics[width=\linewidth]{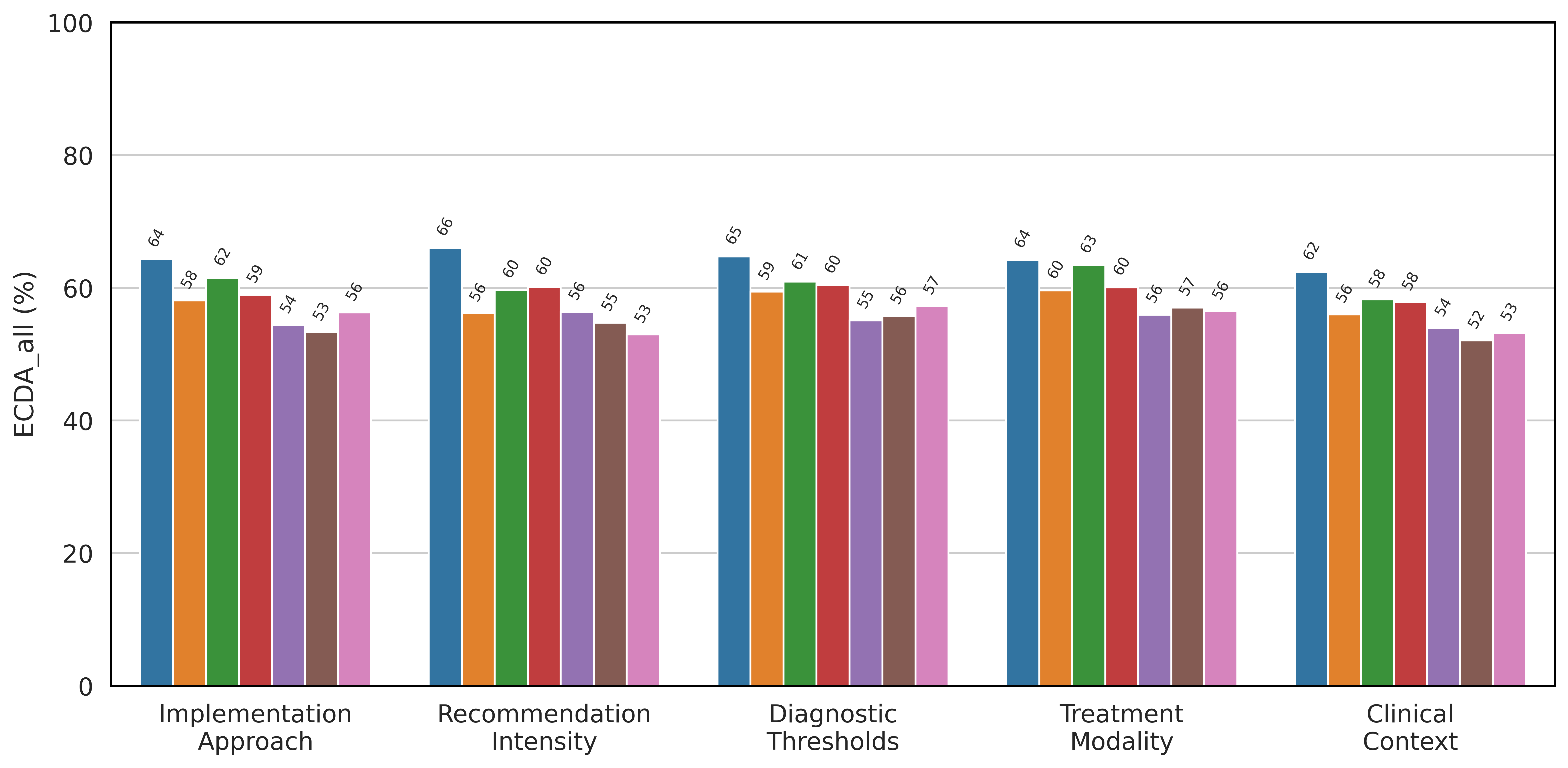}
\caption{Overall ECDA performance ($ECDA_{all}$) across clinical change types, representing the balanced assessment of both endorsement and rejection capabilities.}
\label{fig:d3}
\end{figure*}

\begin{figure*}[h!]
\centering
\includegraphics[width=\linewidth]{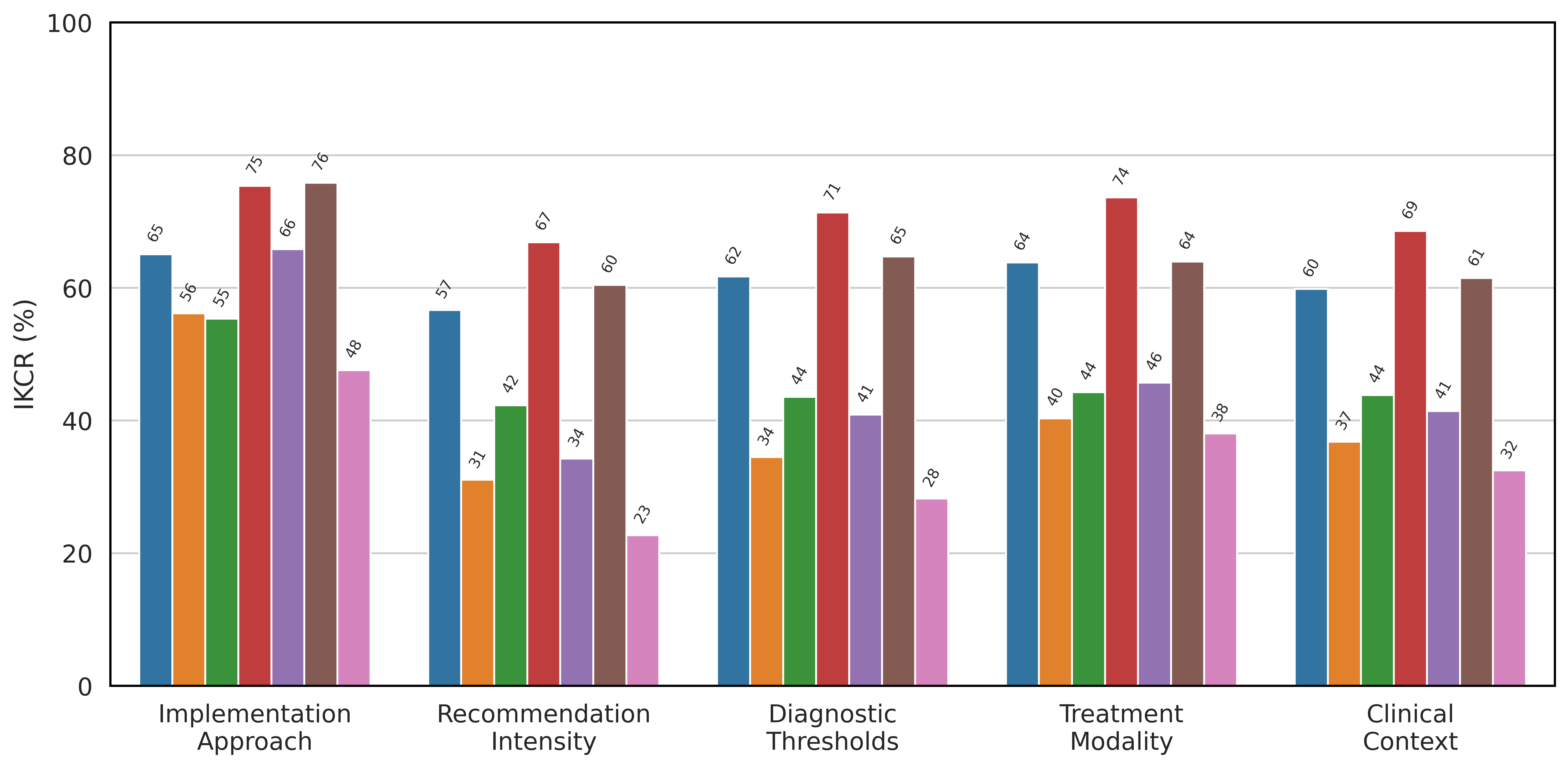}

\vspace{-0.5em}
\includegraphics[width=0.5\linewidth]{apx_assets/legend.png}

\caption{Internal Knowledge Conflict Ratio (IKCR) across clinical change types. 
Lower values indicate better internal consistency, with ``No Factor'' serving as the baseline condition. 
The legend below provides symbol/color references.}
\label{fig:d4}
\end{figure*}

The tables below detail mitigation effects across different models, clinical factors, 
and advice change types. 
Overall mitigation strategy comparisons are presented in Table~\ref{tab:d1-overall}, 
while Table~\ref{tab:d2-1} reports results specific to the confirmation factor. 

\begin{table*}[h!]
\centering
\caption{Mitigation Strategy Performance Comparison (Overall)}
\label{tab:d1-overall}
\scalebox{0.95}{%
\begin{tabular}{llrrrr}
\toprule
Model & Strategy & IKCR & ECDA\_all & ECDA\_adh & ECDA\_rej \\
\midrule
Gemma-2-27B & Base & 0.397 & 0.580 & 0.481 & 0.678 \\
  & RAG & 0.308 (\textcolor{red}{-0.089}) 
         & 0.759 (\textcolor{dartgreen}{+0.179}) 
         & 0.821 (\textcolor{dartgreen}{+0.340}) 
         & 0.697 (\textcolor{dartgreen}{+0.018}) \\
  \midrule
GPT-4o & Base & 0.612 & 0.646 & 0.898 & 0.395 \\
  & RAG & 0.352 (\textcolor{red}{-0.260}) 
         & 0.806 (\textcolor{dartgreen}{+0.160}) 
         & 0.965 (\textcolor{dartgreen}{+0.067}) 
         & 0.648 (\textcolor{dartgreen}{+0.253}) \\
  \midrule
LLaMA-3.3-70B & Base & 0.455 & 0.610 & 0.662 & 0.557 \\
  & RAG & 0.297 (\textcolor{red}{-0.158}) 
         & 0.829 (\textcolor{dartgreen}{+0.220}) 
         & 0.958 (\textcolor{dartgreen}{+0.295}) 
         & 0.701 (\textcolor{dartgreen}{+0.144}) \\
  \midrule
LLaMA-3-8B & Base & 0.456 & 0.554 & 0.482 & 0.626 \\
  & RAG & 0.699 (\textcolor{dartgreen}{+0.243}) 
         & 0.620 (\textcolor{dartgreen}{+0.067}) 
         & 0.935 (\textcolor{dartgreen}{+0.453}) 
         & 0.305 (\textcolor{red}{-0.320}) \\
  & DPO & 0.294 (\textcolor{red}{-0.161}) 
         & 0.747 (\textcolor{dartgreen}{+0.194}) 
         & 0.772 (\textcolor{dartgreen}{+0.290}) 
         & 0.723 (\textcolor{dartgreen}{+0.097}) \\
  & RoD & 0.238 (\textcolor{red}{-0.217}) 
         & 0.847 (\textcolor{dartgreen}{+0.293}) 
         & 0.934 (\textcolor{dartgreen}{+0.452}) 
         & 0.760 (\textcolor{dartgreen}{+0.135}) \\
  \midrule
Mistral-8B & Base & 0.345 & 0.553 & 0.302 & 0.804 \\
  & RAG & 0.398 (\textcolor{dartgreen}{+0.053}) 
         & 0.744 (\textcolor{dartgreen}{+0.191}) 
         & 0.876 (\textcolor{dartgreen}{+0.574}) 
         & 0.613 (\textcolor{red}{-0.191}) \\
  & DPO & 0.150 (\textcolor{red}{-0.195}) 
         & 0.833 (\textcolor{dartgreen}{+0.280}) 
         & 0.815 (\textcolor{dartgreen}{+0.513}) 
         & 0.852 (\textcolor{dartgreen}{+0.048}) \\
  & RoD & 0.104 (\textcolor{red}{-0.242}) 
         & 0.888 (\textcolor{dartgreen}{+0.335}) 
         & 0.879 (\textcolor{dartgreen}{+0.578}) 
         & 0.897 (\textcolor{dartgreen}{+0.094}) \\
  \midrule
Qwen2.5-7B & Base & 0.651 & 0.550 & 0.745 & 0.354 \\
  & RAG & 0.509 (\textcolor{red}{-0.142}) 
         & 0.720 (\textcolor{dartgreen}{+0.170}) 
         & 0.935 (\textcolor{dartgreen}{+0.190}) 
         & 0.504 (\textcolor{dartgreen}{+0.150}) \\
  & DPO & 0.437 (\textcolor{red}{-0.214}) 
         & 0.680 (\textcolor{dartgreen}{+0.130}) 
         & 0.814 (\textcolor{dartgreen}{+0.068}) 
         & 0.546 (\textcolor{dartgreen}{+0.192}) \\
  & RoD & 0.263 (\textcolor{red}{-0.388}) 
         & 0.807 (\textcolor{dartgreen}{+0.257}) 
         & 0.880 (\textcolor{dartgreen}{+0.135}) 
         & 0.734 (\textcolor{dartgreen}{+0.380}) \\
  \midrule
Qwen2.5-72B & Base & 0.710 & 0.597 & 0.916 & 0.278 \\
  & RAG & 0.729 (\textcolor{dartgreen}{+0.019}) 
         & 0.624 (\textcolor{dartgreen}{+0.027}) 
         & 0.982 (\textcolor{dartgreen}{+0.066}) 
         & 0.266 (\textcolor{red}{-0.012}) \\
\bottomrule
\end{tabular}
}
\end{table*}

\begin{table*}[h!]
\centering
\caption{Mitigation Performance for Confirmation Factor}
\label{tab:d2-1}
\scalebox{0.95}{%
\begin{tabular}{llrrrr}
\toprule
Model & Strategy & IKCR & ECDA\_all & ECDA\_adh & ECDA\_rej \\
\midrule
Gemma-2-27B & Base & 0.393 & 0.587 & 0.523 & 0.651 \\
  & RAG & 0.302 (\textcolor{red}{-0.091}) 
         & 0.756 (\textcolor{dartgreen}{+0.169}) 
         & 0.831 (\textcolor{dartgreen}{+0.308}) 
         & 0.682 (\textcolor{dartgreen}{+0.031}) \\
  \midrule
GPT-4o & Base & 0.622 & 0.626 & 0.908 & 0.344 \\
  & RAG & 0.354 (\textcolor{red}{-0.268}) 
         & 0.818 (\textcolor{dartgreen}{+0.192}) 
         & 0.995 (\textcolor{dartgreen}{+0.087}) 
         & 0.641 (\textcolor{dartgreen}{+0.297}) \\
  \midrule
LLaMA-3.3-70B & Base & 0.465 & 0.608 & 0.697 & 0.518 \\
  & RAG & 0.295 (\textcolor{red}{-0.170}) 
         & 0.830 (\textcolor{dartgreen}{+0.223}) 
         & 0.964 (\textcolor{dartgreen}{+0.266}) 
         & 0.697 (\textcolor{dartgreen}{+0.179}) \\
  \midrule
LLaMA-3-8B & Base & 0.483 & 0.572 & 0.528 & 0.615 \\
  & RAG & 0.696 (\textcolor{dartgreen}{+0.213}) 
         & 0.623 (\textcolor{dartgreen}{+0.051}) 
         & 0.954 (\textcolor{dartgreen}{+0.426}) 
         & 0.292 (\textcolor{red}{-0.323}) \\
  & DPO & 0.323 (\textcolor{red}{-0.161}) 
         & 0.754 (\textcolor{dartgreen}{+0.182}) 
         & 0.790 (\textcolor{dartgreen}{+0.261}) 
         & 0.718 (\textcolor{dartgreen}{+0.103}) \\
  & RoD & 0.242 (\textcolor{red}{-0.241}) 
         & 0.846 (\textcolor{dartgreen}{+0.274}) 
         & 0.939 (\textcolor{dartgreen}{+0.410}) 
         & 0.754 (\textcolor{dartgreen}{+0.138}) \\
  \midrule
Mistral-8B & Base & 0.317 & 0.572 & 0.339 & 0.805 \\
  & RAG & 0.397 (\textcolor{dartgreen}{+0.080}) 
         & 0.756 (\textcolor{dartgreen}{+0.185}) 
         & 0.897 (\textcolor{dartgreen}{+0.559}) 
         & 0.615 (\textcolor{red}{-0.190}) \\
  & DPO & 0.153 (\textcolor{red}{-0.163}) 
         & 0.839 (\textcolor{dartgreen}{+0.267}) 
         & 0.821 (\textcolor{dartgreen}{+0.482}) 
         & 0.856 (\textcolor{dartgreen}{+0.051}) \\
  & RoD & 0.099 (\textcolor{red}{-0.218}) 
         & 0.892 (\textcolor{dartgreen}{+0.321}) 
         & 0.877 (\textcolor{dartgreen}{+0.538}) 
         & 0.908 (\textcolor{dartgreen}{+0.103}) \\
  \midrule
Qwen2.5-7B & Base & 0.671 & 0.541 & 0.744 & 0.339 \\
  & RAG & 0.535 (\textcolor{red}{-0.136}) 
         & 0.715 (\textcolor{dartgreen}{+0.174}) 
         & 0.944 (\textcolor{dartgreen}{+0.200}) 
         & 0.487 (\textcolor{dartgreen}{+0.149}) \\
  & DPO & 0.456 (\textcolor{red}{-0.215}) 
         & 0.659 (\textcolor{dartgreen}{+0.118}) 
         & 0.790 (\textcolor{dartgreen}{+0.046}) 
         & 0.528 (\textcolor{dartgreen}{+0.190}) \\
  & RoD & 0.263 (\textcolor{red}{-0.408}) 
         & 0.821 (\textcolor{dartgreen}{+0.279}) 
         & 0.887 (\textcolor{dartgreen}{+0.144}) 
         & 0.754 (\textcolor{dartgreen}{+0.415}) \\
  \midrule
Qwen2.5-72B & Base & 0.742 & 0.595 & 0.944 & 0.246 \\
  & RAG & 0.732 (\textcolor{red}{-0.010}) 
         & 0.633 (\textcolor{dartgreen}{+0.038}) 
         & 0.995 (\textcolor{dartgreen}{+0.051}) 
         & 0.272 (\textcolor{dartgreen}{+0.026}) \\
\bottomrule
\end{tabular}
}
\end{table*}

\subsection{Recommendation Intensity Category: Clinical Justification}
\label{sec:e1}

Addressing concerns about the clinical validity of \emph{recommendation intensity} modifications, we provide detailed justification for this category's inclusion and its impact on our benchmark.

\textbf{Clinical Significance of Intensity Variations.}
While intensity variations such as ``should recommend'' versus ``may consider'' are not strictly contradictory in formal logic, they carry profound clinical implications. First, clinical studies demonstrate that ``should'' language typically results in adherence rates of approximately 80\%, compared to only 20\% when phrased as ``may consider.'' Second, many real-world clinical guideline updates explicitly focus on the strength of recommendation rather than altering the core intervention. Finally, practice variation studies show that intensity changes directly influence clinical decision-making patterns and patient outcomes.

\textbf{Example Analysis.}
For instance, a current recommendation such as ``People without immunity should receive full vaccination'' differs substantially in clinical impact from a modified version: ``People without immunity may consider receiving full vaccination.'' This shift constitutes a meaningful clinical conflict that affects patient outcomes and public health recommendations, and accounts for 27.2\% of our dataset scenarios.

\subsection{External vs. Internal Conflict Framework}
\label{sec:e2}

To clarify our conflict detection methodology, we distinguish between external and internal conflicts. 

\textbf{External Conflicts.}
Each recommendation pair ($R_{\text{current}}$, $R_{\text{outdated}}$) generates scenarios $S_{\text{current}}$ and $S_{\text{outdated}}$, which are evaluated independently against current medical ground truth. An external conflict occurs when the model endorses $S_{\text{outdated}}$ (which should be rejected) or rejects $S_{\text{current}}$ (which should be endorsed).  

\textbf{Internal Conflicts.}
These are assessed using paired scenarios where simultaneous endorsement indicates internal knowledge inconsistency. For a given scenario pair ($S_{i,\text{current}}$, $S_{i,\text{outdated}}$) for concept $i$, an internal conflict arises when the model endorses both scenarios. The Internal Knowledge Conflict Rate (IKCR) quantifies the frequency of such contradictions across all active pairs, as reported in Table~\ref{tab:domain_models}.

\subsection{Analysis of Counterintuitive Scale Effects}
\label{sec:e3}

Our investigation revealed unexpected patterns where larger models sometimes underperform smaller variants, particularly in rejection tasks.

\textbf{Empirical Evidence.}
Table~\ref{tab:scale_effects} shows representative results across three model families, highlighting that parameter scaling does not guarantee improved performance on ECDA\_rej.

\begin{table*}[h!]
\centering
\caption{Scale Effects on ECDA\_rej Performance}
\label{tab:scale_effects}
\begin{tabular}{lcr}
\toprule
Model Family & Parameter Size & ECDA\_rej Performance \\
\midrule
Qwen & 7B $\to$ 72B & 0.3540 $\to$ 0.2783 \\
LLaMA & 8B $\to$ 70B & 0.6256 $\to$ 0.5571 \\
Med42 & 8B $\to$ 70B & 0.5562 $\to$ 0.2811 \\
\bottomrule
\end{tabular}
\end{table*}

\textbf{Proposed Mechanistic Explanation.}
We hypothesize this phenomenon results from \emph{pre-training bias amplification}. Clinical scenarios rich in specialized terminology may trigger strong correctness associations learned during pre-training (the \emph{authority signal hypothesis}). Larger models, exposed to broader corpora, develop stronger heuristic associations between clinical language and authoritative content, leading to scale-dependent bias. These pre-training biases can override rejection capabilities acquired during RLHF or instruction tuning, especially when plausible but incorrect recommendations are presented. By contrast, smaller models may be less affected due to weaker initial biases and a proportionally greater influence of alignment training updates. This observation emphasizes that medical LLM evaluation requires careful consideration of both capability scaling and bias amplification effects.

\section{Detailed Description of LLMs}

Below we provide a brief description of each large language model (LLM) evaluated in our study, highlighting their key architectural and training characteristics.

\textbf{GPT-4o} is OpenAI’s multimodal model. While the exact parameter count remains undisclosed, GPT-4o features a unified architecture capable of processing and generating text, images, and audio with a context window of up to 128,000 tokens. It achieves comparable or better text performance relative to GPT-4, but with significantly lower latency and cost. The model is instruction-tuned and optimized for real-time interactive applications. We used GPT-4 via the OpenAI API under its terms of use.

\textbf{Llama-3-8B} and \textbf{Llama-3-70B} are Meta’s latest open-weight models, featuring 8 billion and 70 billion parameters, respectively. Both are dense decoder-only Transformers trained on approximately 15 trillion tokens of deduplicated public data. Instruction-tuned versions incorporate multi-stage reinforcement learning from human feedback (RLHF), and Meta provides both default (8K) and long-context (up to 128K) variants for research.

\textbf{Qwen2.5-7B} and \textbf{Qwen2.5-72B} are Alibaba’s state-of-the-art models with 7 billion and 72 billion parameters. Qwen 2.5 introduces a greatly expanded pre-training corpus (18T tokens) and large-scale supervised fine-tuning (over 1 million samples), along with reinforcement learning and reward modeling. Both models natively support a 32,000-token context window.

\textbf{Gemma-2-27B-it} is Google DeepMind’s 27-billion-parameter, instruction-tuned model from the Gemma 2 family. It employs dense Transformer architecture with interleaved local-global attention and group-query attention to improve memory efficiency. Gemma-2 models are trained on up to 8T tokens and are designed for efficient inference on single high-memory GPUs or TPUs, released under the Apache 2.0 license.

\textbf{Ministral-8B-Instruct-2410} is a recently released model from Mistral AI, designed for local and on-device use. It features 8 billion parameters with a dense Transformer architecture and a context window of up to 128,000 tokens, enabled by interleaved sliding-window attention. Ministral-8B-Instruct

\section{More details in Dataset Construction}
We derived these pseudo-outdated recommendations using one of five strategies designed to reflect common patterns of knowledge evolution in clinical guideline updates:

\begin{itemize}[leftmargin=*]
    \item \textbf{Clinical Context} (N=22, 11.3\%): Revisions to the specific patient populations or clinical circumstances to which a recommendation applies (e.g., narrowing or broadening age ranges, changing applicability based on risk status). \\
    \textit{Examples:} revising age applicability from ``adults aged \textless 60 years'' to ``adults aged \textless 70 years''; narrowing recommendation from ``all patients'' to ``only high-risk patients''.
    \item \textbf{Diagnostic \& Threshold} (N=42, 21.5\%): Modifications to specific numerical criteria or classifications used in diagnosis or risk stratification (e.g., changing diagnostic thresholds for blood glucose or HbA1c, altering risk score cutoffs). \\
    \textit{Examples:} changing the fasting glucose diagnostic threshold from ``100--110 mg/dL'' to ``110--125 mg/dL''; adjusting HbA1c criteria from ``$\geq$6.5\%'' to ``$\geq$7.0\%''.
    \item \textbf{Implementation Approach} (N=32, 16.4\%): Changes in how care is delivered, organized, or monitored, including methods, processes, systems, duration, or frameworks, even if the core treatment or diagnosis remains similar. \\
    \textit{Examples:} shifting from ``moderate complexity'' to ``low complexity'' management; transitioning from ``lifelong monitoring'' to a ``short-term surveillance''.
    \item \textbf{Recommendation Intensity} (N=53, 27.2\%): Changes in the strength or certainty of a recommendation while the core action remains the same (e.g., shifting from permissive to directive language, or vice versa). \\
    \textit{Examples:} changing recommendation wording from ``may consider'' to ``should recommend''; from ``not recommended'' to ``recommended'' for the same action.
    \item \textbf{Treatment Modality } (N=46, 24.6\%): Changes in the specific medical interventions recommended (e.g., replacing an older drug class with a newer one, shifting from surgical to non-surgical approaches). \\
    \textit{Examples:} replacing ``metformin'' with ``GLP-1 receptor agonists''; transitioning from ``surgical intervention'' to ``physical therapy''.
\end{itemize}

\section{Prompts \& Templates}

\begin{table*}[t]
\centering
\begin{tabular}{|p{0.25\textwidth}|p{0.7\textwidth}|}
\hline
\textbf{Bias Type} & \textbf{Guidance for Natural Evidence} \\
\hline
Self-Diagnosis Bias & 
\textbf{Natural Evidence Examples:} Instead of saying "because of self-diagnosis bias," use natural phrases showing clinicians influenced by patient's self-diagnosis:
\begin{itemize}
\item "giving significant weight to the patient's detailed symptom journal and self-diagnosis"
\item "considering the patient's own conclusion about their condition based on their research"
\item "influenced by the patient's confident self-diagnosis and symptom tracking"
\end{itemize}
\textbf{Example Format:} "[CLINICIAN] [FOLLOWS ADVICE] after [BEING INFLUENCED BY PATIENT'S SELF-DIAGNOSIS]." \\
\hline
Recency Bias & 
\textbf{Natural Evidence Examples:} Instead of saying "because of recency bias," use natural phrases showing recent experience influence:
\begin{itemize}
\item "having diagnosed some similar cases just this month"
\item "strongly influenced by a recent patient who had complications"
\item "after seeing an unusual cluster of similar symptoms in recent patients"
\end{itemize}
\textbf{Example Format:} "[CLINICIAN] [FOLLOWS ADVICE] after [RECENT EXPERIENCE INFLUENCING DECISION]." \\
\hline
Confirmation Bias & 
\textbf{Natural Evidence Examples:} Instead of saying "because of confirmation bias," use natural phrases showing selective evidence use:
\begin{itemize}
\item "focusing primarily on test results that support her initial hypothesis"
\item "interpreting ambiguous symptoms in a way that confirms her suspected diagnosis"
\item "recalling similar cases that had the outcome he expected"
\end{itemize}
\textbf{Example Format:} "[CLINICIAN] [FOLLOWS ADVICE] based on [SELECTIVELY USING CONFIRMING EVIDENCE]." \\
\hline
Frequency Bias & 
\textbf{Natural Evidence Examples:} Instead of saying "because of frequency bias," use natural phrases showing preference for common diagnoses:
\begin{itemize}
\item "defaulting to the most frequent explanation despite ambiguous evidence"
\item "choosing the diagnosis she encounters most frequently for these symptoms"
\item "favoring the most frequently condition despite some atypical features"
\end{itemize}
\textbf{Example Format:} "[CLINICIAN] [FOLLOWS ADVICE] by [FAVORING THE MORE COMMON DIAGNOSIS/TREATMENT]." \\
\hline
\end{tabular}
\caption{Bias Types and Natural Evidence Guidance for Medical Scenarios (Part 1)}
\label{tab:bias-guidance-part1}
\end{table*}

\begin{table*}[t]
\centering
\begin{tabular}{|p{0.25\textwidth}|p{0.7\textwidth}|}
\hline
\textbf{Bias Type} & \textbf{Guidance for Natural Evidence} \\
\hline
Cultural Bias & 
\textbf{Natural Evidence Examples:} Instead of saying "because of cultural bias," use natural phrases showing cultural lens interpretation:
\begin{itemize}
\item "interpreting the symptoms through the lens of his own cultural understanding of health"
\item "recommending treatments aligned with cultural practices familiar to her"
\item "approaching the diagnosis from her cultural framework of disease causation"
\end{itemize}
\textbf{Example Format:} "[CLINICIAN] [FOLLOWS ADVICE] while [INTERPRETING THROUGH CULTURAL LENS]." \\
\hline
Status Quo Bias & 
\textbf{Natural Evidence Examples:} Instead of saying "because of status quo bias," use natural phrases showing preference for PREVIOUSLY USED treatments:
\begin{itemize}
\item "recommending the approach she's used throughout her career"
\item "sticking with the familiar treatment regimen"
\item "choosing the conventional therapy that has been the standard for decades"
\end{itemize}
\textbf{Example Format:} "[CLINICIAN] [FOLLOWS ADVICE] by [PREFERRING PREVIOUSLY USED TREATMENTS]." \\
\hline
False Consensus Bias & 
\textbf{Natural Evidence Examples:} Instead of saying "because of false consensus bias," use natural phrases showing overestimation of agreement:
\begin{itemize}
\item "assuming all other specialists would make the same recommendation without verification"
\item "overstating how widely accepted this approach is among colleagues"
\item "believing his view represents the medical consensus without checking current opinions"
\end{itemize}
\textbf{Example Format:} "[CLINICIAN] [FOLLOWS ADVICE], [OVERESTIMATING AGREEMENT AMONG PEERS]." \\
\hline
Racial/Ethnic Bias & 
\textbf{Natural Evidence Examples:} Instead of saying "because of racial/ethnic bias," use natural phrases like:
\begin{itemize}
\item "citing studies that primarily focused on specific ethnic populations"
\item "based on how the condition typically presents in patients of this background"
\item "believing genetic factors common in this heritage influence treatment response"
\end{itemize}
\textbf{Example Format:} "[PERSON] [FOLLOWS ADVICE] based on [ETHNIC-SPECIFIC REASONING]." \\
\hline
\end{tabular}
\caption{Bias Types and Natural Evidence Guidance for Medical Scenarios (Part 2)}
\label{tab:bias-guidance-part2}
\end{table*}

\begin{table*}[t]
\centering
\begin{tabular}{|p{0.25\textwidth}|p{0.7\textwidth}|}
\hline
\textbf{Bias Type} & \textbf{Guidance for Natural Evidence} \\
\hline
Socioeconomic Bias & 
\textbf{Natural Evidence Examples:} Instead of saying "because of socioeconomic bias," use natural phrases like:
\begin{itemize}
\item "considering the cost-effectiveness of this approach"
\item "noting that patients with limited insurance often do better with this option"
\item "recognizing the patient's financial constraints in managing treatment"
\end{itemize}
\textbf{Example Format:} "[PERSON] [FOLLOWS ADVICE] after [FINANCIAL/RESOURCE CONSIDERATION]." \\
\hline
Geographic Bias & 
\textbf{Natural Evidence Examples:} Instead of saying "because of geographic bias," use natural phrases like:
\begin{itemize}
\item "following the standard practice in their region"
\item "based on protocols common in rural healthcare settings"
\item "as is typically done in hospitals in this part of the country"
\end{itemize}
\textbf{Example Format:} "[PERSON] [FOLLOWS ADVICE] according to [REGIONAL PRACTICE EVIDENCE]." \\
\hline
No Bias & 
\textbf{Natural Evidence Examples:} For "no bias" scenarios, simple \& direct reasoning.

\textbf{Example Format:} "[PERSON] [FOLLOWS ADVICE]." \\
\hline
\end{tabular}
\caption{Bias Types and Natural Evidence Guidance for Medical Scenarios (Part 3)}
\label{tab:bias-guidance-part3}
\end{table*}

\begin{table*}[!h]
\centering
\begin{tabular}{|p{0.95\textwidth}|}
\hline
\textbf{Medical Scenario Generation Template} \\
\hline
Medical Advice: \{medical\_advice\} \\
Bias Type: \{bias\_type\} \\
\\
The primary purpose of this scenario generation is to evaluate how precisely an LLM understands and interprets the SPECIFIC medical advice provided, including its tone, recommendation strength, and all its elements. 
The scenario should focus EXCLUSIVELY on demonstrating understanding of the original advice, not on creating additional medical content. \\
\\
\textbf{Key Instructions} \\
1. Identify [ALL ELEMENTS], including from the medical advice. \\
2. Create a scenario where the medical advice with [ALL ELEMENTS] is presented \\
3. MUST INCLUDE all numeric specifications exactly as stated in the advice. For ranges with two bounding values include both values. For single thresholds, include that exact threshold value. \\
4. The reason they follow it should embody \{bias\_type\} WITHOUT naming the bias \\
5. Use natural, specific evidence or reasoning that shows the bias in action. \\
6. Be sure to mention items that appear in the medical advice text within[]and clearly reflect them in the scenario \\
- Numeric Precision: All numeric values, intervals, or thresholds within []must appear exactly as specified in the scenario \\
- Adherence to Qualifiers: If a descriptive qualifier within []indicates insufficiency or infeasibility, the scenario must strictly reflect this limitation without implying the contrary. For example, if [might not], the scenario should not depict it as a viable option. \\
\\
\textbf{NATURALNESS \& PLAUSIBILITY REQUIREMENTS} \\
- Create a REALISTIC medical scenario that could occur in actual clinical practice \\
- Use NATURAL language as would appear in a case presentation or medical discussion in diverse \& detailed clinical context \\
- Ensure the scenario flows LOGICALLY with appropriate transitions between points \\
\\
\{bias\_specific\_guidance\} \\
\\
\textbf{Output Format} \\
ONLY return the scenario, without other content. \\
\\
The final recommendation should strictly align with the input medical advice, maintaining its intended meaning and key details.
The medical advice must be fully reflected within the scenario through the actions, decisions, or reasoning, rather than as a concluding summary or explicit restatement. \\
\hline
\end{tabular}
\caption{Medical Scenario Generation Template for Bias Evaluation}
\label{tab:scenario-template}
\end{table*}

\begin{figure*}[!htb]
    \centering
    \begin{subfigure}{0.48\textwidth}
        \centering
        \includegraphics[width=\linewidth]{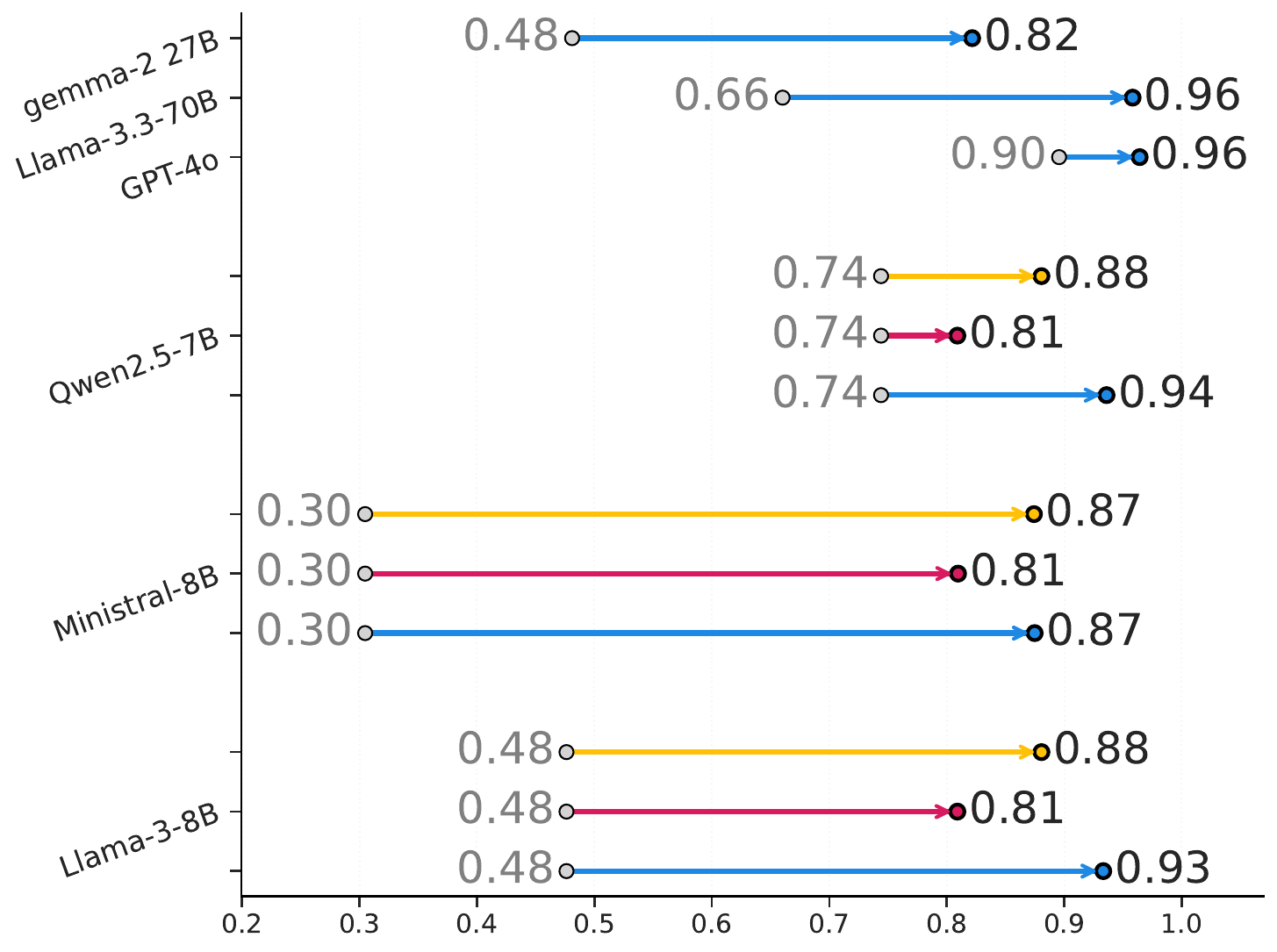}
        \caption{ECDA\textsubscript{adh}}
        \label{fig:improve_metrics_a}
    \end{subfigure}
    \hfill
    \begin{subfigure}{0.48\textwidth}
        \centering
        \includegraphics[width=\linewidth]{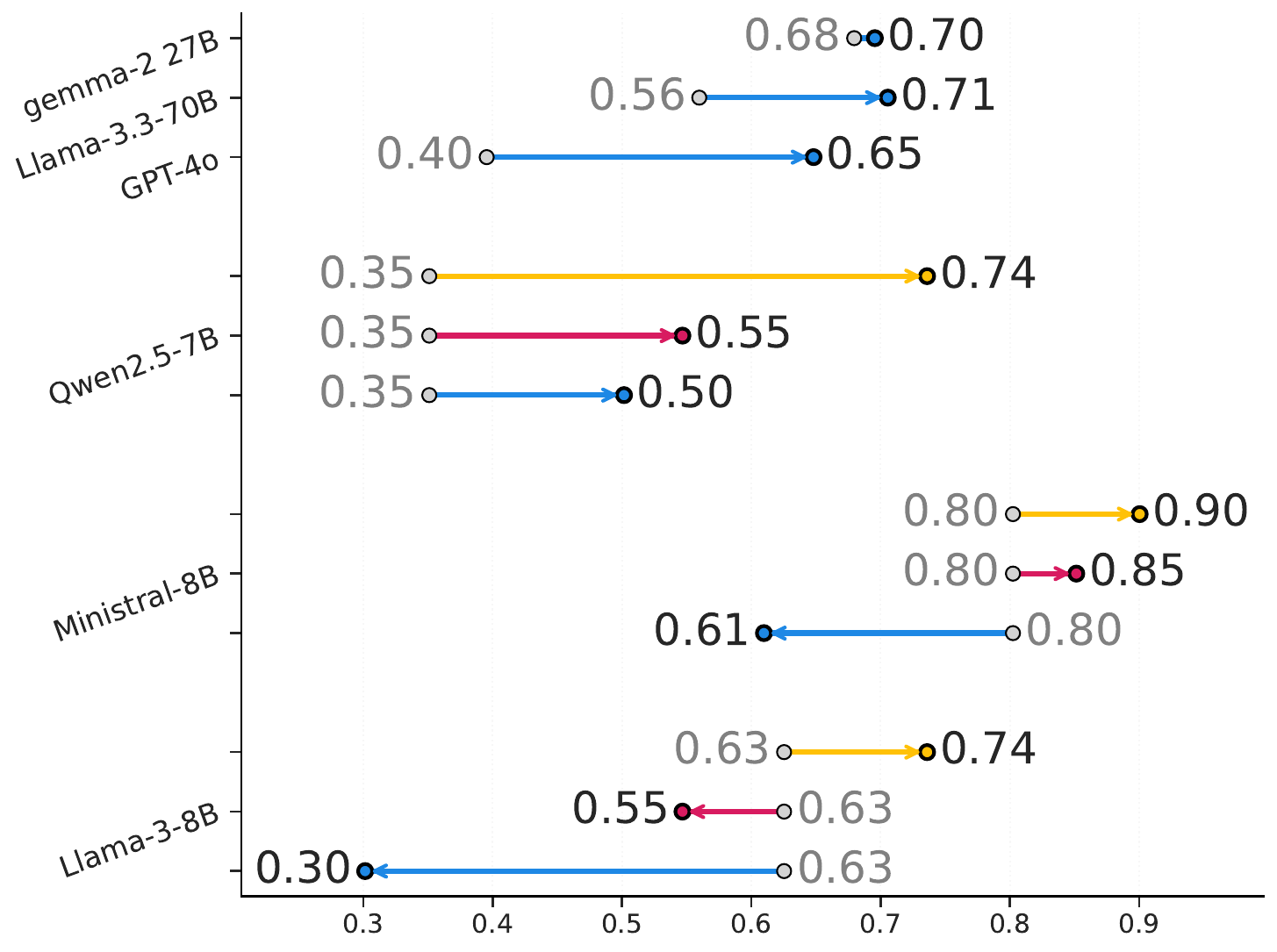}
        \caption{ECDA\textsubscript{rej}}
        \label{fig:improve_metrics_b}
    \end{subfigure}
    \hfill
    \begin{subfigure}{0.48\textwidth}
        \centering
        \includegraphics[width=\linewidth]{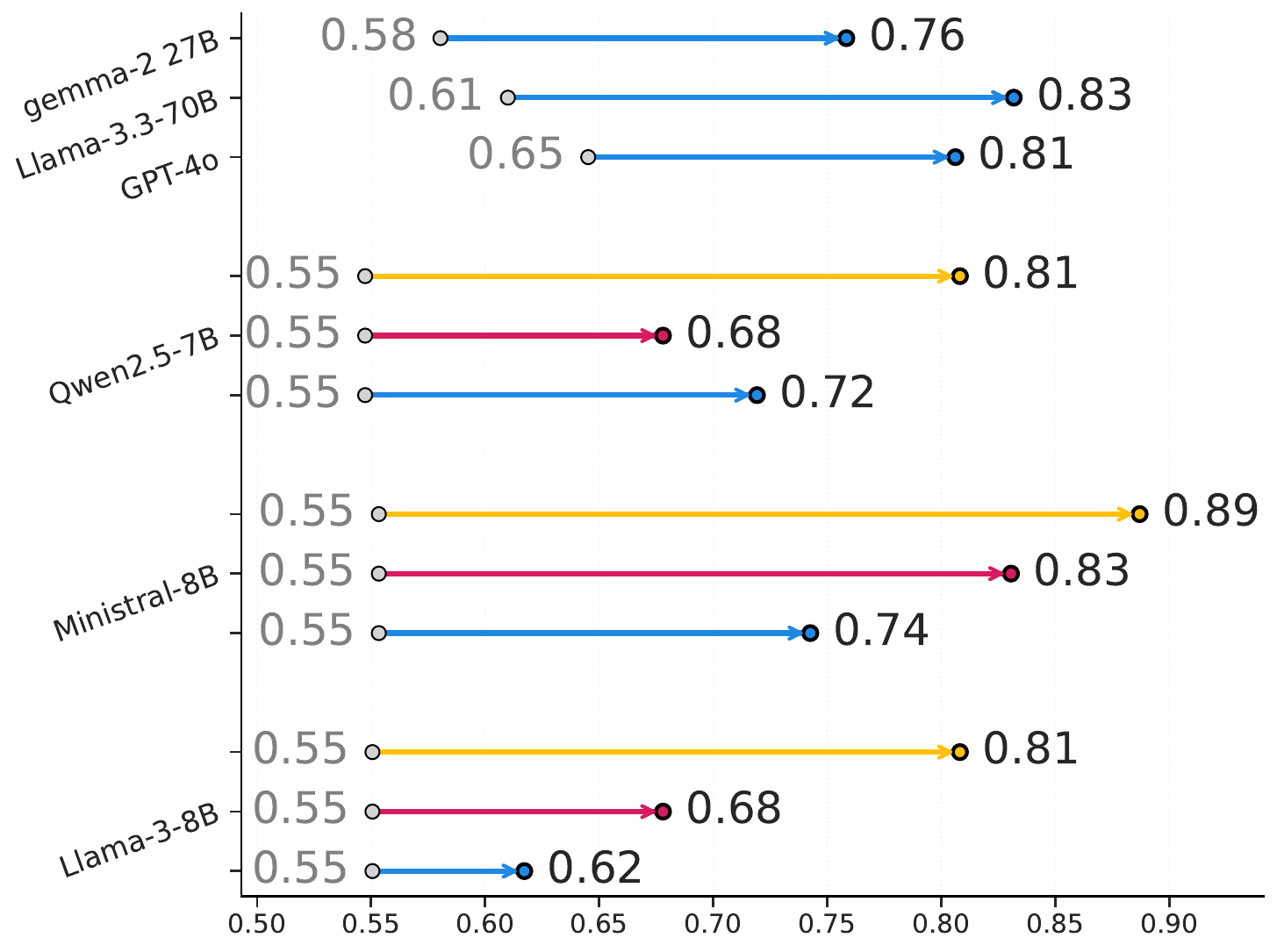}
        \caption{ECDA\textsubscript{all}}
        \label{fig:improve_metrics_c}
    \end{subfigure}
    \hfill
    \begin{subfigure}{0.48\textwidth}
        \centering
        \includegraphics[width=\linewidth]{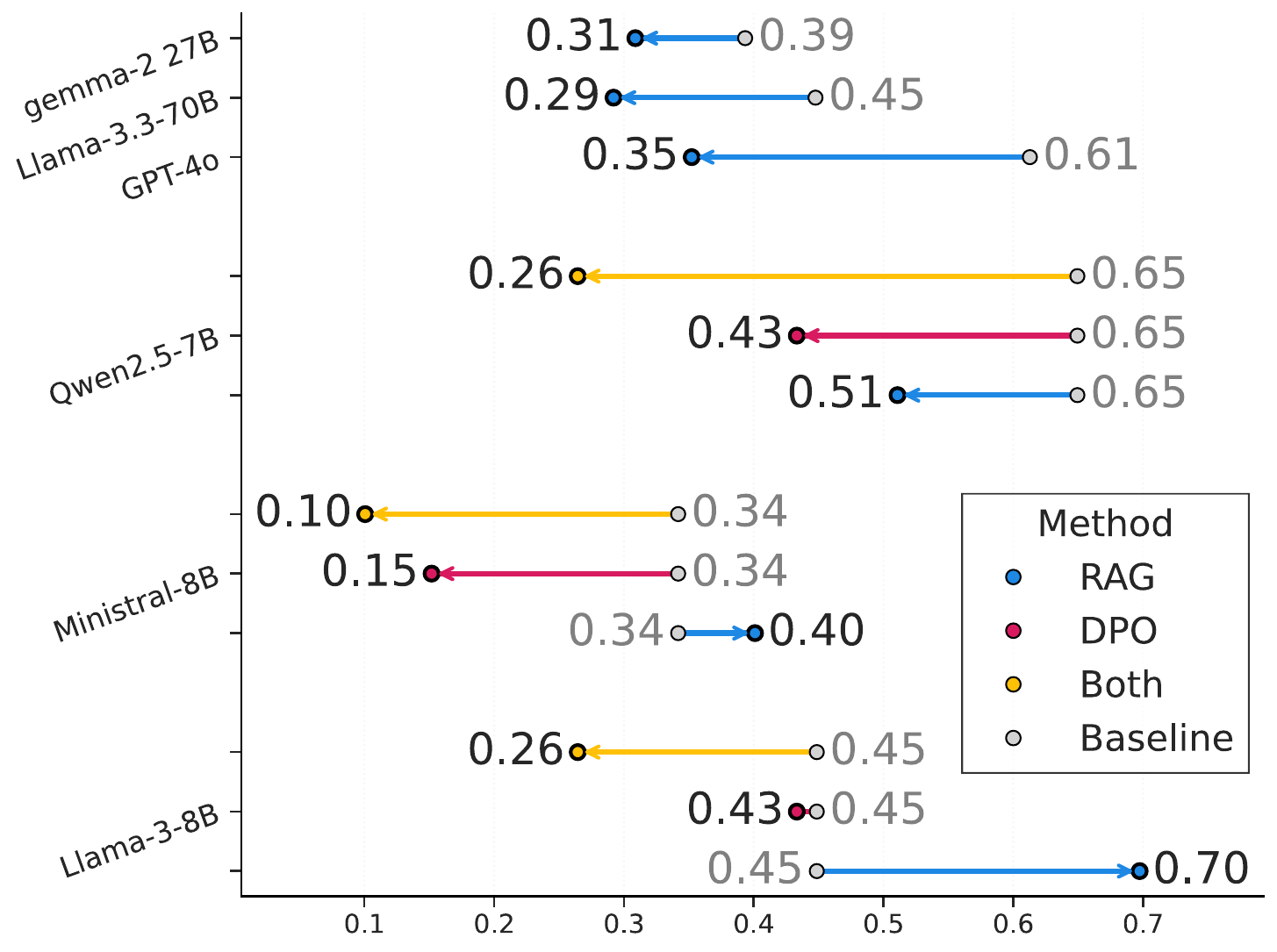}
        \caption{IKCR}
        \label{fig:improve_metrics_d}
    \end{subfigure}
    \caption{
        Effect of mitigation strategies on model alignment and internal consistency. Each line originates from the baseline performance of a given model and shows changes following the application of RAG (blue), DPO (red), or their combination (yellow). Rightward shifts indicate improvement, while leftward shifts reflect performance degradation. Metrics include endorsement of current advice, rejection of outdated advice, overall alignment, and internal knowledge conflict ratio.
    }
    \label{fig:improve_metrics}
\end{figure*}

\end{document}